\documentclass[sigconf]{acmart}

\usepackage{microtype}
\usepackage{graphicx}
\usepackage{booktabs} 
\usepackage{subcaption}
\usepackage{makecell}
\usepackage{hyperref}

\usepackage{algorithm}
\usepackage{algpseudocode}

\usepackage{amsmath}
\usepackage{amsfonts}
\usepackage{mathtools}
\usepackage{amsthm}
\usepackage{xcolor,colortbl}
\usepackage{multirow}
\usepackage{fancyhdr}

\DeclareMathOperator*{\PGD}{\it PGD}
\usepackage{bbm}
\usepackage{multirow}
\usepackage{bbding}
\usepackage{enumitem}

\theoremstyle{plain}
\newtheorem{theorem}{Theorem}[section]

\newtheorem{lemma}[theorem]{Lemma}

\theoremstyle{definition}

\theoremstyle{remark}

\usepackage{xspace} 
\newcommand{\ie}{\emph{i.e.,}\xspace}
\newcommand{\eg}{\emph{e.g.,}\xspace}

\copyrightyear{2024}
\acmYear{2024}
\setcopyright{rightsretained}
\acmConference[KDD '24]{Proceedings of the 30th ACM SIGKDD Conference on
Knowledge Discovery and Data Mining}{August 25--29, 2024}{Barcelona, Spain}
\acmBooktitle{Proceedings of the 30th ACM SIGKDD Conference on Knowledge
Discovery and Data Mining (KDD '24), August 25--29, 2024, Barcelona,
Spain}\acmDOI{10.1145/3637528.3672009}
\acmISBN{979-8-4007-0490-1/24/08}

\begin{document}

\title[RC-NAS for Versatile Adversarial Robustness]{Reinforced Compressive Neural Architecture Search for Versatile Adversarial Robustness}

\author{Dingrong Wang}
\email{dw7445@rit.edu}
\affiliation{%
  \institution{Rochester Institute of Technology}
  \city{Rochester}
  \state{NY}
  \country{USA}
}
\author{Hitesh Sapkota}
\authornote{Work does not relate to position at Amazon.}
\email{sapkoh@amazon.com}
\affiliation{%
  \institution{Amazon Inc.
  }
  \city{Sunnyvale}
  \state{CA}
  \country{USA}
}
\author{Zhiqiang Tao}
\email{zhiqiang.tao@rit.edu}
\affiliation{
  \institution{Rochester Institute of Technology}
  \city{Rochester}
  \state{NY}
  \country{USA}
}
\author{Qi Yu}
\authornote{Correspondence author.}
\email{qi.yu@rit.edu}
\affiliation{%
  \institution{Rochester Institute of Technology}
  \city{Rochester}
  \state{NY}
  \country{USA}
}

\begin{abstract}
Prior neural architecture search (NAS) for adversarial robustness works have discovered that a lightweight and adversarially robust neural network architecture could exist in a non-robust large teacher network, generally disclosed by heuristic rules through statistical analysis and neural architecture search, generally disclosed by heuristic rules from neural architecture search.  
However, heuristic methods cannot uniformly handle different adversarial attacks and "teacher" network capacity. To solve this challenge, we propose a Reinforced Compressive Neural Architecture Search (RC-NAS) for Versatile Adversarial Robustness. Specifically, we define task settings that compose datasets, adversarial attacks, and teacher network information. Given diverse tasks, we conduct a novel dual-level training paradigm that consists of a meta-training and a fine-tuning phase to effectively expose the RL agent to diverse attack scenarios (in meta-training), and making it adapt quickly to locate a sub-network (in fine-tuning) for any previously unseen scenarios. Experiments show that our framework could achieve adaptive compression towards different initial teacher networks, datasets, and adversarial attacks, resulting in more lightweight and adversarially robust architectures.
\end{abstract}




\maketitle

\section{Introduction}
\label{sec:intro}
Deep neural networks (DNNs) have benefited many real-world applications, such as image classification \cite{krizhevsky2009learning}, object detection \cite{lin2014microsoft}, and natural language processing \cite{rajpurkar2016squad}. However, standard DNNs are vulnerable to adversarial attacks, raising an effective remedy to include deeper and/or wider blocks along with adversarial training~\cite{zhang2019theoretically, madry2017towards, Wang2020Improving, wu2020adversarial, rade2022reducing}. Since such strategies may incur significant computational overhead, recent efforts have been devoted to locating lightweight architectures that are robust to different adversarial attacks through neural architecture search (NAS)~\cite{devaguptapu2021, zhu2023improving, huang2021exploring,huang2022revisiting,guo2020meets, sehwag2020hydra}. 

\begin{figure*}
    \centering
    \begin{subfigure}{0.24\linewidth}
      \centering
      \includegraphics[width=\linewidth]{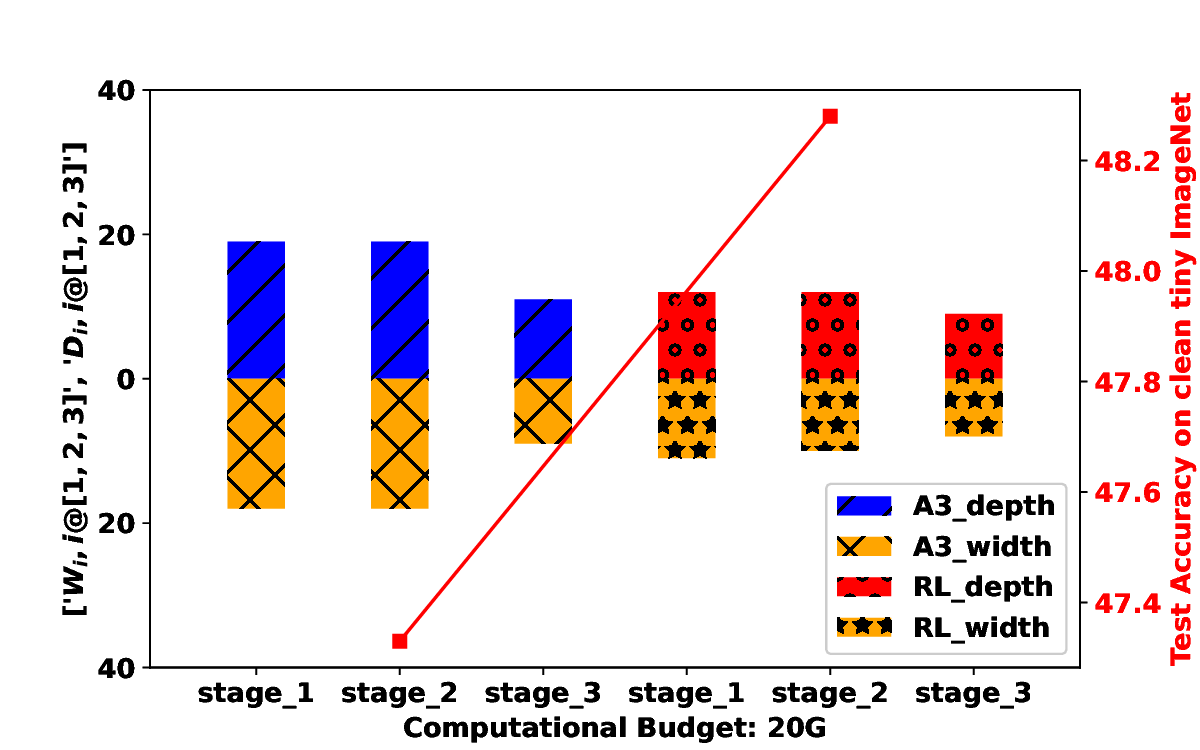}
      \caption{\label{fig:tiny_20g_clean} 20G on clean }
    \end{subfigure}
    \begin{subfigure}{0.24\linewidth}
      \centering
      \includegraphics[width=\linewidth]{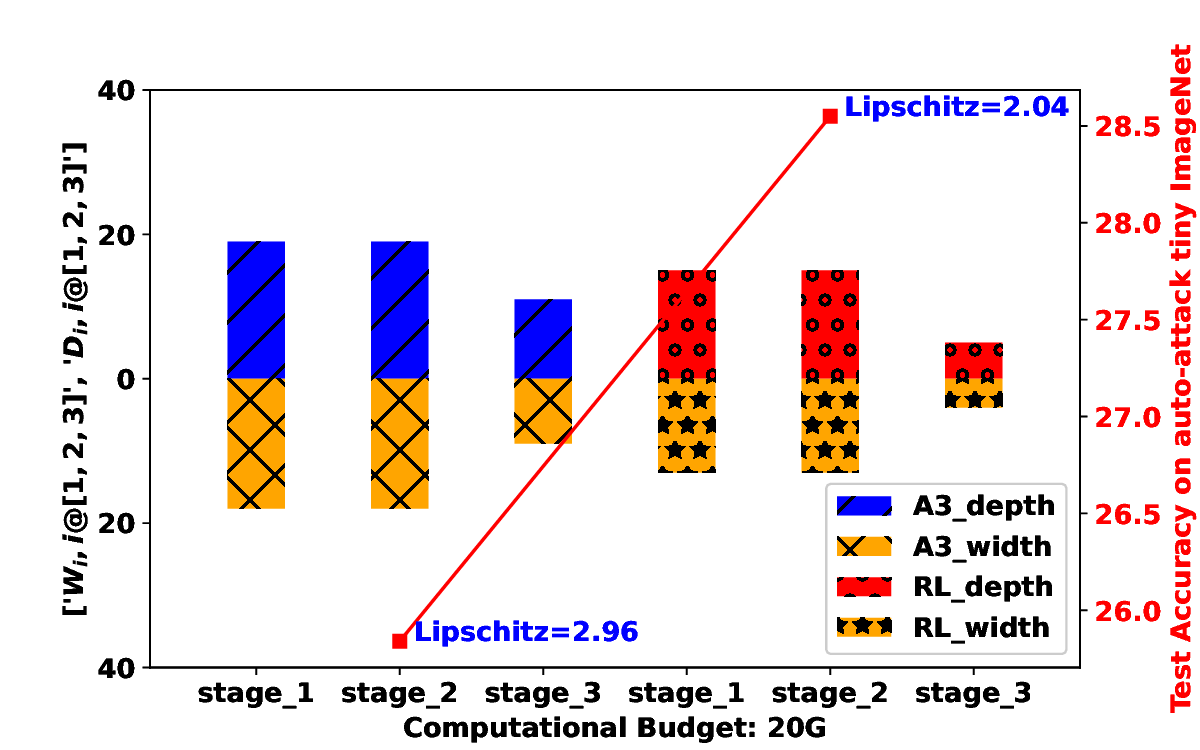}
      \caption{\label{fig:tiny_20g_auto} 20G on auto-attack }
    \end{subfigure}
    \begin{subfigure}{0.24\linewidth}
      \centering
      \includegraphics[width=\linewidth]{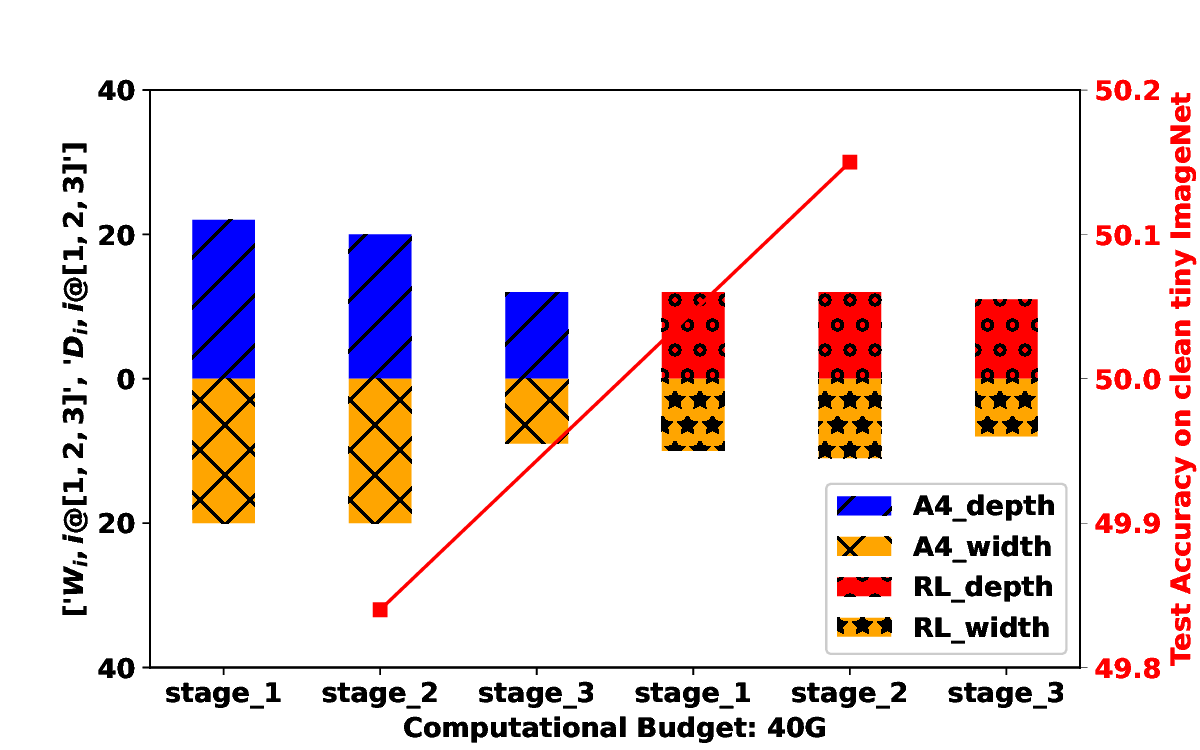}
      \caption{\label{fig:tiny_40g_clean} 40G on clean }
    \end{subfigure}
    \begin{subfigure}{0.24\linewidth}
      \centering
      \includegraphics[width=\linewidth]{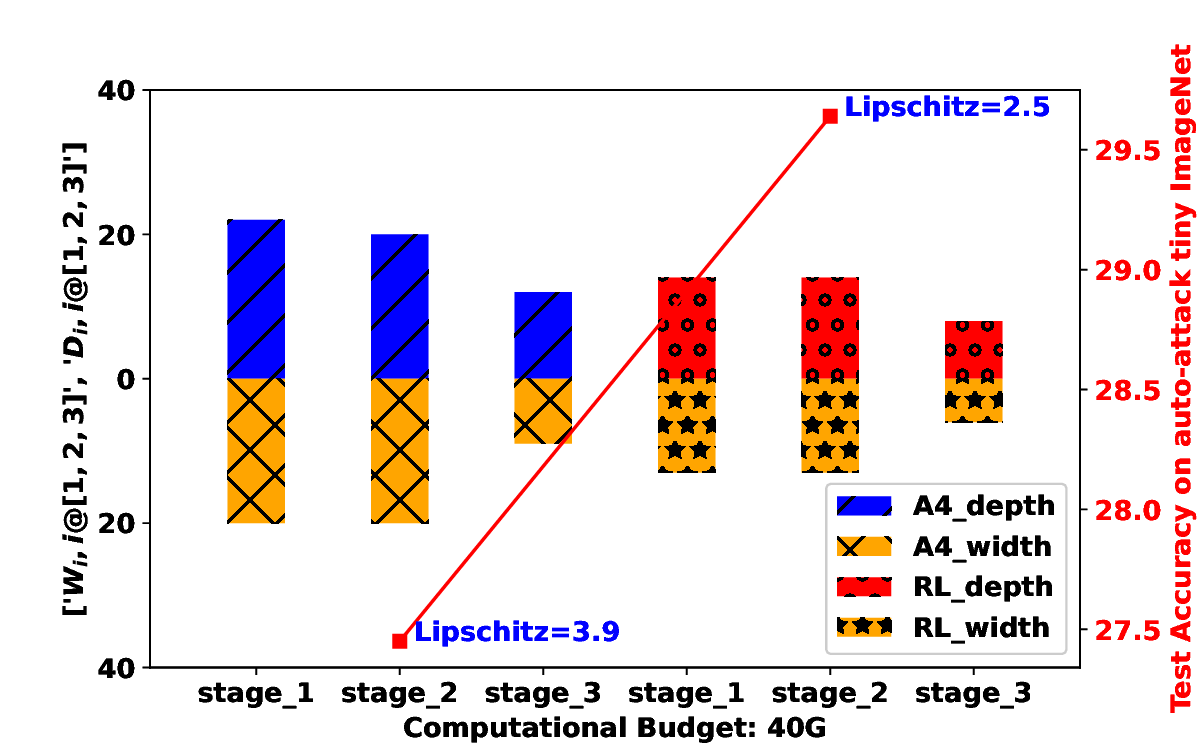}
      \caption{\label{fig:tiny_40g_auto} 40G on auto-attack }
    \end{subfigure}\\
    \begin{subfigure}{0.24\linewidth}
      \centering
      \includegraphics[width=\linewidth]{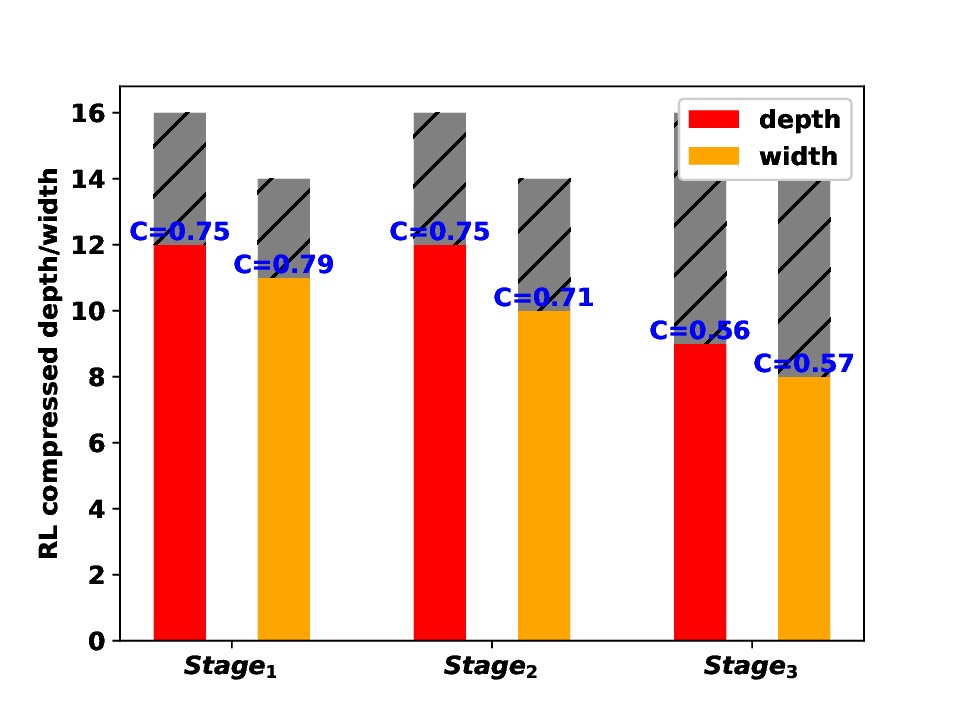}
      \caption{\label{fig:tiny_20g_clean_rl} WRN-46-14 on clean }
    \end{subfigure}
    \begin{subfigure}{0.24\linewidth}
      \centering
      \includegraphics[width=\linewidth]{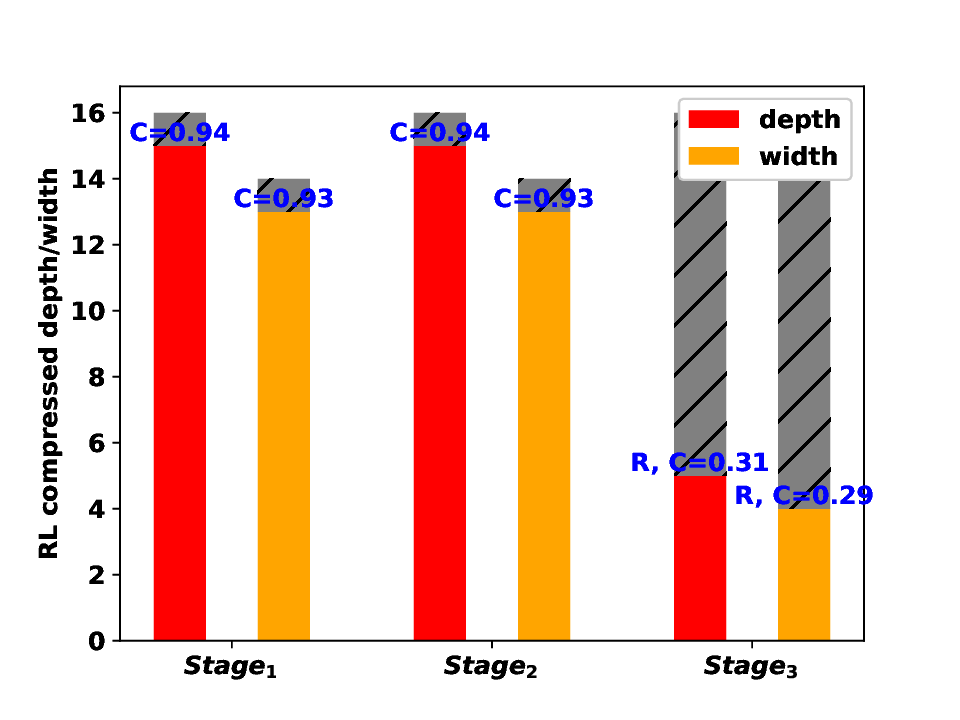}
      \caption{\label{fig:tiny_20g_auto_rl} WRN-46-14 on auto-attack }
    \end{subfigure}
    \begin{subfigure}{0.24\linewidth}
      \centering
      \includegraphics[width=\linewidth]{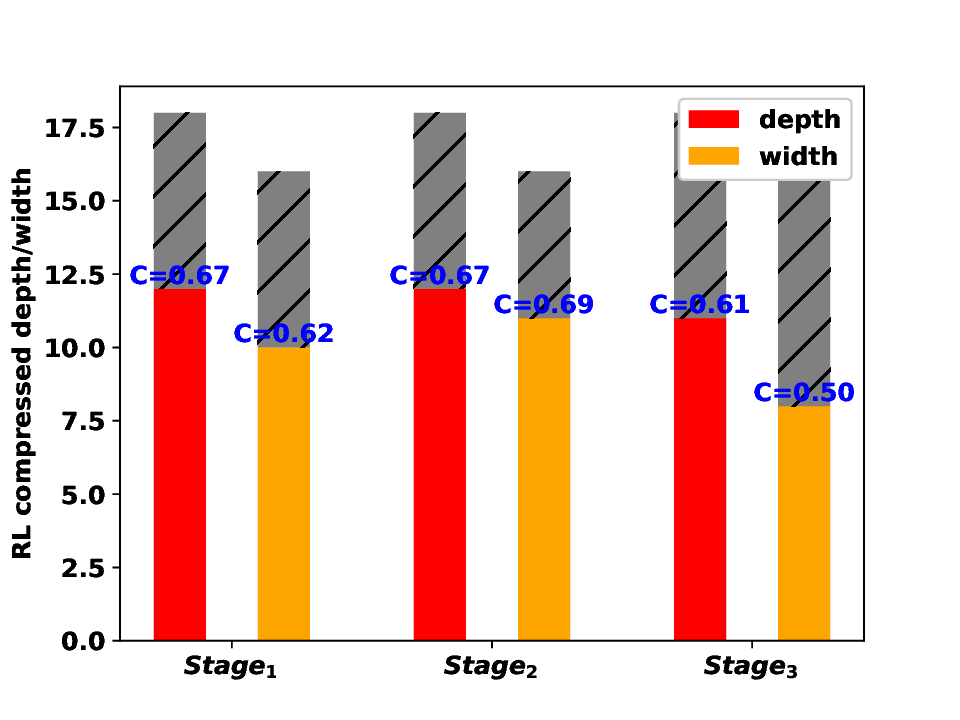}
      \caption{\label{fig:tiny_40g_clean_rl} WRN-70-16 on clean }
    \end{subfigure}
    \begin{subfigure}{0.24\linewidth}
      \centering
      \includegraphics[width=\linewidth]{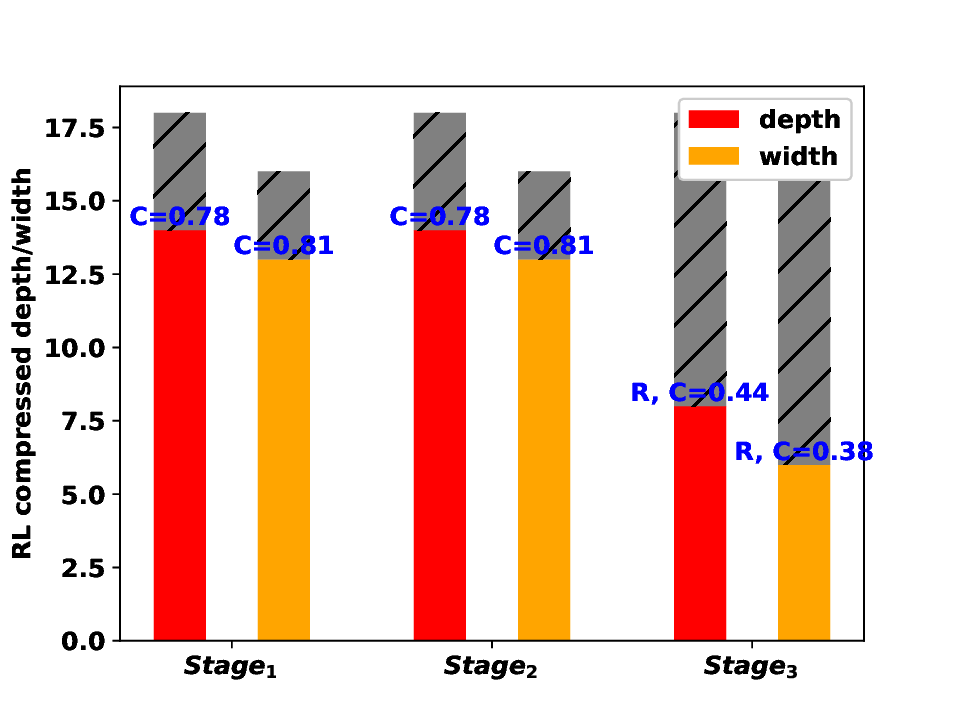}
      \caption{\label{fig:tiny_40g_auto_rl} WRN-70-16 on auto-attack }
    \end{subfigure}
    \vspace{-2mm}
    \caption{Architecture topology analysis of \texttt{RobustResNet} in different attack scenarios from the mildest (\ie clean) to the most severe (\ie auto-attack) on Tiny-ImageNet. WRN-46-14 and WRN-70-16 are leveraged as teacher networks with 20 and 40 GFLOPs computation budgets, respectively. The corresponding student networks are referred to as \texttt{RobustResNet-A3} and \texttt{RobustResNet-A4}. The entire teacher network architecture is partitioned into multiple (\eg 3) stages as in \cite{huang2022revisiting} and we visualize both depths and widths of the corresponding \texttt{RobustResNet} for each stage. In (a)-(d), the left three bar plots (in blue and orange) show the depths and widths in the three stages of \texttt{RobustResNet} A3 and A4 that follow the same configuration rules; the right three bar plots (in red and orange) show the adaptive configuration obtained by the proposed reinforced learning (RL) based architecture search. 
    In (e)-(h), the grey bars denote the capacity of the corresponding teacher networks and $C$ denotes the remaining percentage of each stage after compression.
    }
    \label{fig:result_tiny}
    \vspace{-2mm}
\end{figure*}

One mainstream direction in NAS is to leverage a generative process to seek the best-performed network architectures based on a manually designed library of architectural ingredients. However, given the complex nature of DNN architectures coupled with the diverse types of adversarial attacks, such a process can become too costly to cover various aspects, making it hard to guarantee a good adversarially robust performance. Another line of work suggests that there exists an optimal architectural configuration for adversarial robustness in a large non-robust ``teacher'' architecture, which enjoys a smaller parameter size and better robustness \cite{huang2021exploring, sehwag2020hydra}. Consequently, network-to-network (N2N) compression could be conducted to achieve adversarial robustness and has shown promising progress in recent works~\cite{huang2021exploring,huang2022revisiting}. For example, Huang et al.~\cite{huang2022revisiting} investigated over 1,000 network architectures randomly sampled from some large teacher networks and selected top-ranked robust sub-networks. Empirically, they derived a set of useful rules that can help to guide the design of robust ResNet (\texttt{RobustResNet}) architectures from different teacher networks and computation budgets, varying from 5G to 40G. However, most of these rules are derived in a heuristic way, offering no guarantee of achieving an optimal trade-off between compression ratio and adversarial accuracy as the learning environment changes, which thus may lead to suboptimal performance. As shown in Figure~\ref{fig:result_tiny}, \texttt{RobustResNet}~\cite{huang2022revisiting} suffers from a lower adversarial accuracy due to adopting the fixed configuration rules across different attacks and computation budgets.

To overcome the key limitations of existing solutions, we propose a novel reinforcement learning (RL) framework, referred to as Reinforced Compressive Neural Architecture Search (\texttt{RC-NAS}), that leverages the flexibility of a specially designed RL agent supported by a powerful dual-level training paradigm to perform a systematic search over a rich and complex space of architecture configurations. The trained RL agent can quickly adapt to the highly diverse attack scenarios and locate a compressed student sub-network with guaranteed adversarial robustness while meeting the computational budget constraints. The ability to automatically adjust to distinct attack scenarios and adaptively compress the teacher network in different ways (instead of following fixed rules) is a critical step towards realizing truly versatile adversarial robustness that significantly advances the state of the art.

To achieve versatile adversarial robustness, it is essential to identify the key characteristics from different attack scenarios and train the RL agent to recognize them and perform adaptive N2N compression given a specific attack setting. To this end, the proposed RL framework collectively considers important features of the teacher network, the nature of the dataset, the level of adversarial attack, and the overall computational budget. Based on different attack conditions as RL states, the compression ratio of each stage (RL actions) should be adaptive to achieve the best compression and robust accuracy trade-off. Specifically, given an RL state, the agent performs an action to shrink the network by compressing the width and depth at the stage level as well as determining specific configurations (\eg convolution type, activation, and normalization) at the block level. A specially designed award function that integrates adversarial accuracy with computational budget constraint serves as the feedback to encourage the RL to seek for architectures with an optimal robust accuracy and compression trade-off. 


For the RL training, we introduce a novel dual-level training paradigm that consists of a meta-training and a fine-tuning phases to effectively expose the RL agent to diverse attack scenarios so that it can quickly adapt to a sub-network customized towards the unknown attacks during the test phase. In the meta-training phase, a pool of diverse attack tasks with distinct characteristics is formulated. The RL agent is iteratively trained by sampling different tasks from the pool so that it can gain the capability to recognize key patterns from these different combinations of task key factors and perform adaptive N2N compression. In the second phase, the agent is given a more specific attack scenario and it performs fine-tuning by leveraging the meta-trained model as the starting point to achieve quick and effective adaptation. Figure~\ref{fig:result_tiny} (a)-(d) shows that the proposed RL framework can adapt to different attack scenarios by finding highly customized robust sub-networks, instead of relying on a fixed set of configuration rules as in existing works. As a result, it leads to much improved adversarial accuracy with a better compression ratio. Furthermore, as shown in (e)-(h), highly distinct compression ratios are applied to different stages of the teacher network and an overall higher compression ratio is applied to a larger teacher network. More interestingly, the compression ratios also vary dramatically based on the level of attacks: for a more severe attack (\ie auto-attack), the first two stages are compressed much less (to capture the subtle changes in the input) while the last stage is compressed more significantly (to reduce the perturbation caused by the attack). Our theoretical analysis provides further insights on the compression behavior.

To the best of our knowledge, this is the first effort to provide a principled RL framework for searching an adversarially robust architecture in a non-robust large teacher network (network-to-network compression for adversarial robustness). Our main contributions are summarized below:
\begin{itemize}[nosep,leftmargin=4mm]
\item Novel Reinforced Compressive Neural Architecture Search (\texttt{RC-NAS}) framework that leverages the flexibility of reinforcement learning to explore a rich and complex space of architectures for lightweight and adversarially robust sub-networks,
\item Simulated RL environment equipped with a specially designed state encoder and an award function, allowing the RL agent to encode key ingredients from the teacher architecture, the dataset, level of adversarial attack, and the computational budget,
\item Dual-level RL training to enable quick adaption to specific attack settings by exposing the RL agent to diverse attack scenarios, 
\item Deeper theoretical analysis that reveals why the RL driven N2N compression can lead to improved adversarial robustness. 
\end{itemize}

We conduct extensive experiments to demonstrate the effectiveness of the proposed \texttt{RC-NAS} framework over different input conditions, including different teacher network architecture with pairing computation budgets (5G--40G), and visual learning tasks of varying difficulty (CIFAR-10, CIFAR-100, and Tiny-ImageNet). For the adversarially trained RL compressed network, we compare it with the latest N2N compression baselines for adversarial robustness and show its superior performance across different teacher network capacities, datasets, and adversarial attack test conditions. We also emphasize that such good performance is attributed to the novel RL framework and its unique dual-level training paradigm, which is verified by multiple ablative studies and statistical analysis of RL selected robust sub-network architectures.


\vspace{-2mm}\section{Related Work}
\label{sec:related}

\vspace{2mm}\noindent{\bf Neural Architecture Search (NAS).}
There has been much work on exploring the rich design space of neural network architectures \cite{iandola2016squeezenet, zoph2016neural, devaguptapu2021, zhu2023improving}. The principal aim of previous work in architecture search has been to build models that maximize performance on a specific dataset. There has been increasing interest in conduct a N2N style architecture search to achieve adversarial robustness \cite{huang2021exploring,huang2022revisiting,guo2020meets}. For example, Wu et al. theoretically analyze why wider networks tend to have worse perturbation stability on linearized neural tangent kernels and develop a width adjusted regularization (WAR) algorithm to address that \cite{wu2021wider}. 
Huang et al.  emphasize that a higher model capacity does not necessarily improve adversarial robustness, especially in the last stage and there exists an optimal architectural configuration for adversarial robustness under the same parameter budget \cite{huang2021exploring}. To this end, residual networks have been intensively analyzed by considering architecture design at both the block level and the network scaling level \cite{huang2022revisiting}. A robust residual block and a compounding scaling rule have been derived to properly distribute depth and width at the desired computation budget.
However, those proposed heuristic rules for N2N compression are either too general or too specific for diverse adversarial learning scenarios. A flexible way to achieve a robust architecture that can adapt to the unique characteristics of each attack scenario is in demand to achieve truly versatile adversarial robustness, which is the focus of our work.  


\vspace{2mm}\noindent{\bf Network pruning.}
Pruning-based methods preserve the weights that matter most and remove the redundant weights \cite{guo2016dynamic,LeCun1989,frankle2018lottery}. While pruning-based approaches typically operate on the weights of the model, our approach operates on a much larger search space over both model weights and model architecture. Recently some works have combined adversarial learning with network pruning such as Hydra~\cite{sehwag2020hydra} and ADMM~\cite{ye2019adversarial}, but none of them have paid attention to architecture search. Instead, our work focus on reinforced neural architecture search for adversarial robustness, thus providing more flexible architectural design choices and enjoying a compressed parameter space at the same time.

\vspace{2mm}\noindent{\bf Reinforcement learning.}
Deep Reinforcement Learning has been extensively applied in game agent training \cite{mnih2013playing}, natural language contextual understanding \cite{stiennon2020learning}, causal relationship inference \cite{zhang2020causal}, image classification \cite{hafiz2022image}, object detection \cite{Bellver_2016_NIPSWS}, time series analysis \cite{Wang2023DTS} and information retrieval \cite{Wang2022DARP}. In the NAS domain, there are some works \cite{zoph2016neural, ashok2018nn} focusing on designing an RL agent to sample a sub-network within a pre-defined architecture search space. For example, Ashok et al. perform student-teacher knowledge distillation for clean accuracy on small datasets \cite{ashok2018nn}. However, none of existing efforts pay attention to the relationship between the adversarial robustness and the reinforced network-to-network compression, which is main design focus of our RL framework.

\vspace{-2mm}\section{Methodology}
\label{sec:methodology}

\noindent\textbf{Overview.} The overall goal of the proposed \texttt{RC-NAS} framework is to train a RL agent so that it can perform adaptive N2N compression of a large teacher network to achieve lightweight sub-networks robust to specific adversarial attacks. Given a different attack scenario, the RL agent needs to recognize its key characteristics and formulates a customized policy to generated compression actions. To this end, we propose a dual-level training paradigm and employ a meta-training phase to expose the RL agent to diverse attack scenarios. For the action design, we want our RL agent to control both macro stage-level width/depth configuration and micro block-level details, such as convolution types, activation and normalization. To support RL training, we define a formal Markov Decision Process that provides the key building blocks of the RL environment. 



\begin{figure*}
    \centering
    \begin{subfigure}{0.49\linewidth}
        \includegraphics[width=\linewidth]{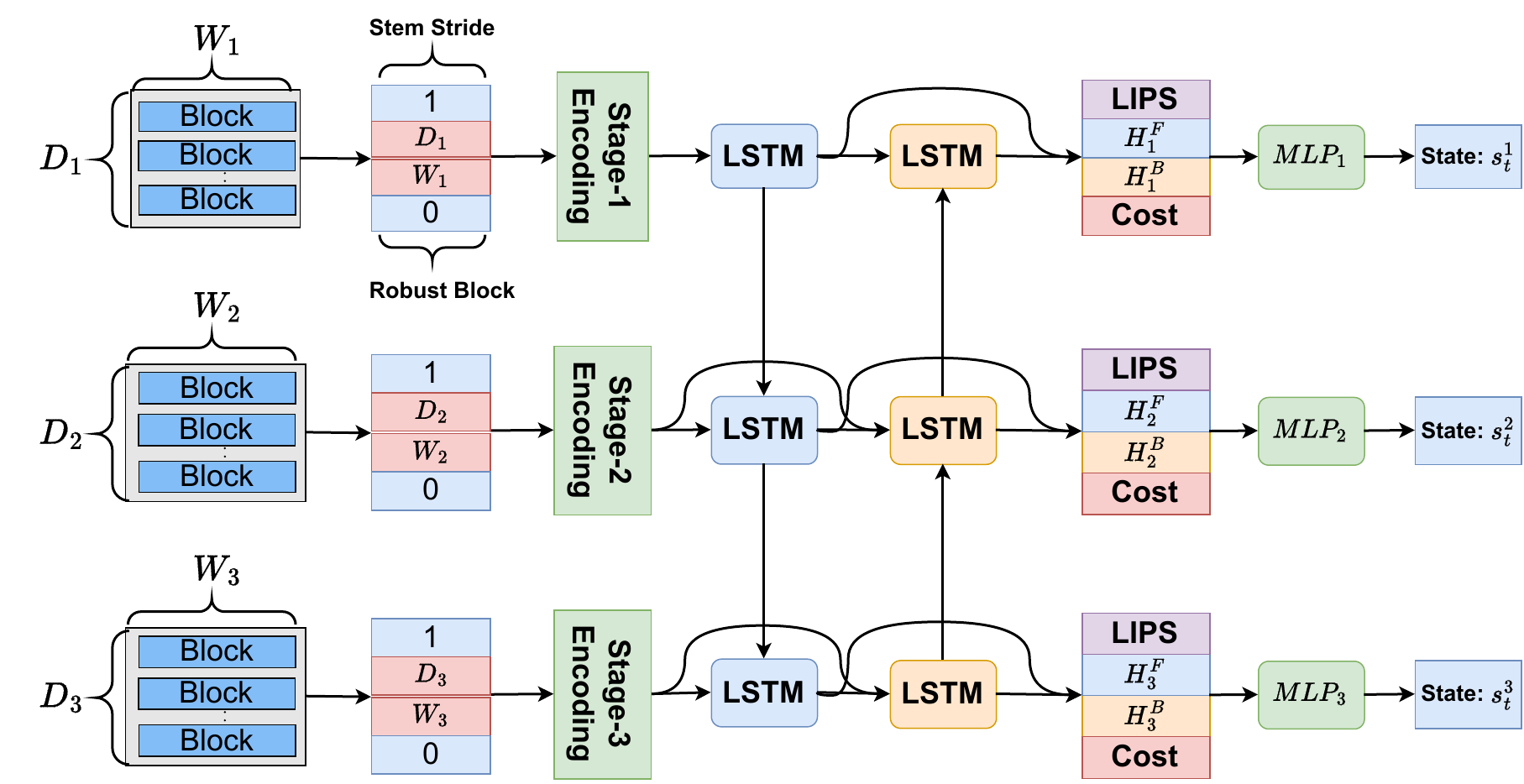}
        \caption{\label{fig:SE}State Encoder}
    \end{subfigure}
    \hfill
    \begin{subfigure}{0.49\linewidth}
        \includegraphics[width=\linewidth]{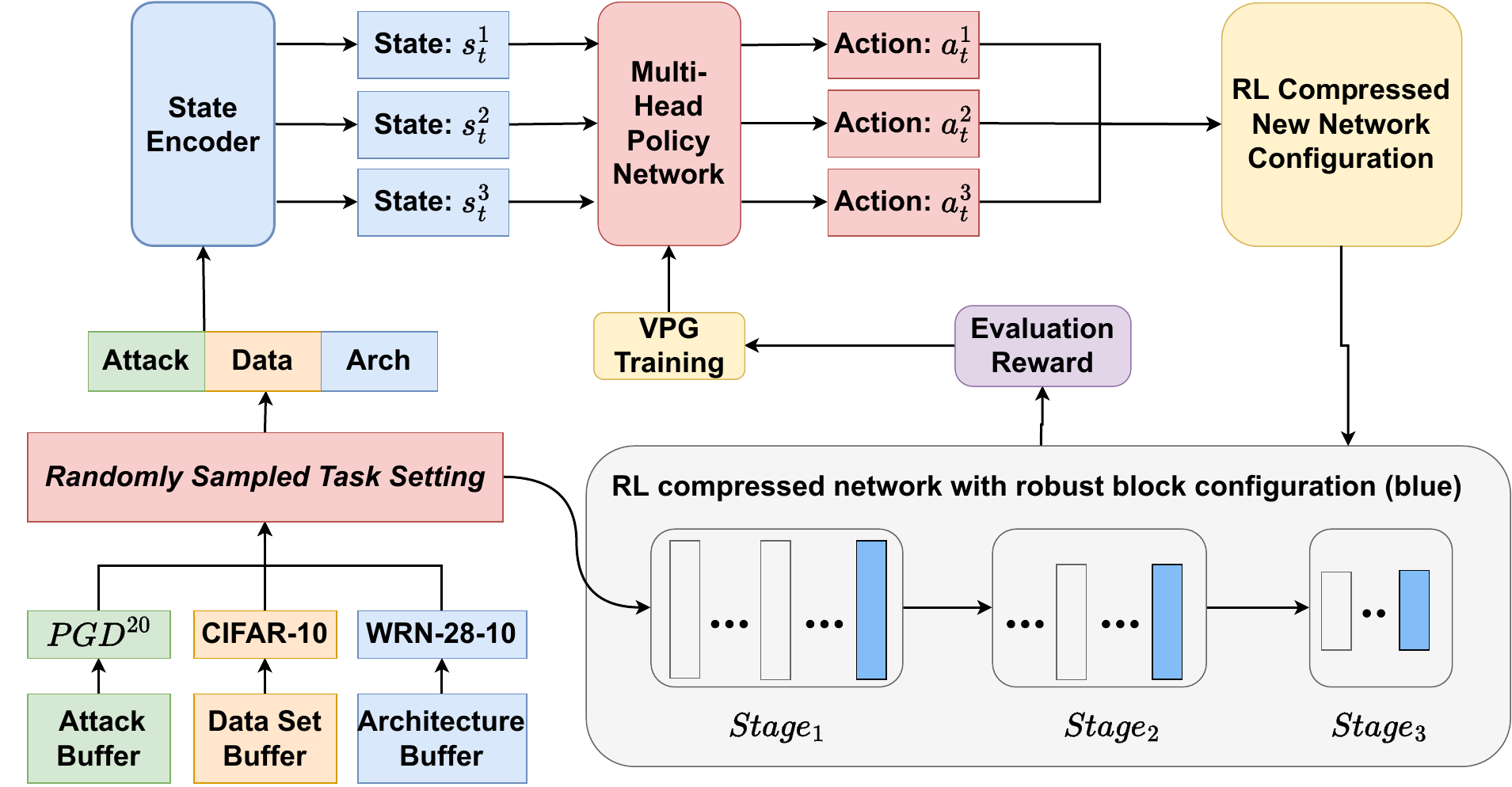}
        \caption{\label{fig:RL_train} RL training framework}
    \end{subfigure}
    \caption{\label{fig:RL}Overview of the \texttt{RC-NAS} framework}
\end{figure*}

\vspace{-2mm}
\subsection{Markov Decision Process}
The Markov Decision Process (MDP) for our proposed \texttt{RC-NAS} defines its own state, action, reward and state transition function:
\begin{itemize}[nosep,leftmargin=4mm]
    \item \textbf{State}: State $s_t$ is an embedding encoding the input teacher network topology, adversarial attack level, dataset complexity, and corresponding computation costs.
    \item \textbf{Action}: Action $a_t$ is an embedding designating RL compressed stage-level and block-level configurations given the input architecture for the current time step.
    \item \textbf{Reward}: Reward $r_t$ is the trained RL-compressed sub-network's evaluation accuracy on a separate evaluation set of the same difficulty as input training set, weighted by compression ratios and normalized by the initial teacher network's performance. Also we consider the desired computation budget and set an annealing penalty if it is not satisfied.
    \item \textbf{State Transition Function:} Transition function $s_{t+1}\sim \mathcal{T}(\cdot|s_t,a_t)$ is achieved by multiple buffers, i.e. an \textit{architecture} buffer which stores all the candidate teacher architectures to compress, a \textit{Data} buffer containing small/medium/large dataset choices, and an \textit{Attack} buffer consisting of different adversarial attack methods. For each time step, we randomly sample one architecture-dataset-attack combination from these buffers as the new input task setting for state encoding.
\end{itemize}

In each time step, we sample the input teacher architecture, adversarial attack, as well as a dataset and encode them into the next state $s_{t+1}$ by leveraging the state encoder. Then, the multi-head policy network takes $s_{t+1}$ to generate a compression action $a_{t+1}$ that specifies the compression operations, such as the remaining percentage of depth and width for each stage, as well as the application of robust block configurations. After getting the newly sampled sub-network, it is included  into the \textit{architecture} buffer and the RL agent moves to the next step. For RL training, we train the RL sampled sub-network on the training set with standard adversarial training and test its adversarial performance on corresponding evaluation set of the same input task setting. The reward is calculated based on the adversarial accuracy normalized by the robust accuracy from the teacher network, and weighted by the compression ratio comparing to the teacher network. We also consider an annealing penalization if the computation budget is not satisfied. The whole RL framework is trained by vanilla policy gradient based on the above defined reward. We introduce the detailed RL settings in the following section. 

\vspace{-2mm}\subsection{Reinforced Neural Architecture}

\vspace{2mm}\noindent\textbf{Attack task formulation}
An attack task is formed by combining an adversarial attack method $\mathcal{P}$, a dataset $D$, and an initial teacher architecture $f_{\psi}$ sampled their corresponding buffers.   
An evaluation set $D_{eval}$ is further separated from the dataset for reward evaluation (we assign 256 data samples to the evaluation set in our experiments). 
After the RL agent is fully trained, it will sample a sub-network from the target task's teacher network $f_{\psi}$, and then the selected sub-network will be trained using a standard adversarial training process on the clean training set $D_{train}$ and evaluated on $\mathcal{P}$-attacked test set $D_{test}$ from the target task.

\vspace{2mm}\noindent\textbf{State encoder.}
As shown in Figure \ref{fig:SE}, the state encoder takes the initial teacher network topology as input, including each stage's width (scalar), depth (scalar), stride step of its first layer indicating whether to down-sample the input feature map into half from the last stage output (binary), and the application of robust block \cite{huang2022revisiting} inside each stage (binary). We split above information into stage encodings which summarize each stage's related information. These stage encodings $\texttt{ST}_{i}, \forall i\in [1,N_{stage}]$ will go through a bi-directional LSTM to output forward and backward encodings $H_i^F$ and $H_i^B$, respectively. The forward encoding summarizes all previous stage information while the backward encoding considers the afterward stage information beyond the current stage. 

To capture the level of adversarial attack, we propose to evaluate the Lipschitz coefficient~\cite{virmaux2018lipschitz} of the attacked dataset with respect to the teacher network. This can be achieved through three steps: 1) Train the teacher network using a standard adversarial training process on the training set. 2) Perform adversarial attack on the clean trained teacher network with evaluation set data applying the adversarial attack method from the same input task setting. 3) Calculate the Lipschitz coefficient of each adversarial attacked data instance on the evaluation set by comparing to the network output from the same clean instance, 
\begin{equation}
\label{eq:lipschitz}
\texttt{LIPS} = \frac{\Vert f_{\psi}(\mathcal{P}(x_i;\hat{\epsilon}))-f_{\psi}(x_i)\Vert_{1}}{\Vert\mathcal{P}(x_i;\hat{\epsilon})-x_i\Vert_\infty}, x_i \in D_{eval}
\end{equation}
where $f_\psi$ is the adversarially trained teacher network, $\mathcal{P}$ is the chosen adversarial attack, and $D_{eval}$ is the evaluation set.

Finally, the capacity of the current input teacher network is represented by its inference speed cost on every data instance of the evaluation set, which is measured by GFLOPs, resulting into another embedding vector \texttt{CT}. By concatenating all four embedding vectors, we generate the state embedding $\texttt{concat}(\texttt{LIPS}, H_i^F, H_i^B, \texttt{CT})$. Through a multi-layer perception module $\texttt{MLP}_i$ we transform the a state embedding into the RL state $s^i_t$,
\begin{align}
\label{eq:state}
s_t^i = \texttt{MLP}_{i}\left(\texttt{concat}(\texttt{LIPS}, H_i^F, H_i^B, \texttt{CT})\right) 
\end{align}

\vspace{1mm}\noindent\textbf{Multi-head policy network.}
Multi-head policy network $f_\phi(\cdot)$ functions as the RL actor to generate actions given state $s_t = \texttt{concat}\{s^i_t, i\in[1,N_{stage}]\}$. The generated action $a_t = \texttt{concat}\{a^i_t, i\in[1,N_{stage}]\}$ where each $a^i_t$ corresponding to $stage_i$ is a four dimensional vector and designates four different compression operations for each stage. The policy network is multi-head after a shared feature extraction module $\mathcal{E}$. Specifically, the our heads are to generate the two pairs of means: $\mu_i=(\mu_{i,1},\mu_{i,2})^\top$ and a covariance matrix $\Sigma_i = \left[ {\begin{array}{cc}
    \sigma_{i,1}^2 & 0 \\
    0 & \sigma_{i,2}^2 \\
  \end{array} } \right]$ for constructing a two-dimensional Gaussian distribution $\mathcal{N}(\mu_i, \Sigma_i)$, to sample the remaining percentage of the width and depth after RL compression. We use a diagonal co-variance matrix $\Sigma_i$ by assuming that width and depth are independent to each other. Another two heads are used to generate the probabilities $p=(p_{i,1},p_{i,2})$ for sampling two binary signals through a multi-Bernoulli distribution $Ber$ to designate whether to half down-sample the input feature map in the stage's first layer through the first binary signal and whether to apply the robust block configuration in that stage through the other binary signal. For those Gaussian and Bernoulli heads with parameters $\phi_{\mathcal{N}}$ and $\phi_{\mathcal{B}}$, we train them with vanilla policy gradient using the reparametrization trick~\cite{kingma2013auto}, and the gradients from two heads will both trace back to and update the shared feature extraction module parameters $\phi_{\mathcal{E}}$. We formulate this process below: 
\begin{align}
\label{eq:action}
& (\mu_i,\Sigma_i) = f_{\phi_{\mathcal{E}}, \phi_{\mathcal{N}}}(s_t^i),\quad p_i = f_{\phi_{\mathcal{E}}, \phi_{\mathcal{B}}}(s_t^i)  \nonumber\\
& a_t^{i} = \texttt{concat}(\sigma(\alpha_i), \beta_i), \alpha_i \sim \mathcal{N}(\cdot|\mu^i,\Sigma^i), \beta_i \sim Ber(\cdot|p_i)
\end{align}
where $\sigma$ is the sigmoid activation function.


\vspace{2mm}\noindent\textbf{RL reward design.}
The evaluation reward is defined in Eq.~\eqref{eq:reward}. First, the generated network compressed by the RL actions is trained on the training set of the given task using standard adversarial training, and then evaluated on the adversarially attacked evaluation set of the task to get the adversarial accuracy $\Tilde{A}_{RL}$. Then, the accuracy is normalized by the adversarially trained teacher network's evaluation accuracy $\Tilde{A}_{teacher}$ and then weighted by a compression ratio $C$ comparing to the teacher network, we use the quadratic form $C(2-C)$ to smoothly encourage a higher compression due to the characteristics of the quadratic curve, where $C=1$ reaches the peak of curve $C(2-C)$. We also consider the desired computation budget $CB$ which is measured by GFLOPs. If the average inference speed (GFLOPs) of the RL compressed sub-network on evaluation set $\zeta_{RL}=Average(Cost)$ surpasses this computation budget, the reward will be penalized by an annealing factor $\epsilon$ which starts 1 and gradually reduce to 0 as training goes. Therefore in the beginning of training, RL agent is encouraged to find the best adversarial robust architectures suitable to the task setting while not paying too much attention to the computation budget limit, and as it is trained to grasp the necessary knowledge about architecture choices, it will try to fulfill the computation budget requirements while following its summarized rules for compressing the network.
\begin{align}
\label{eq:reward}
r(s_t,a_t)= 
\begin{cases}
  C(2-C) \cdot\frac{\Tilde{A}_{RL}}{\Tilde{A}_{teacher}} & \zeta_{RL} \leq CB \\
  \epsilon\left(C(2-C) \cdot\frac{\Tilde{A}_{RL}}{\Tilde{A}_{teacher}}+1\right)-1  & otherwise
\end{cases}
\end{align}

\noindent\textbf{Optimization.}
The state encoder is pre-trained separately as Figure~\ref{fig:pretrain} shows in Appendix~\ref{app:pre_train}. After its pre-training, the state encoder's weight will be frozen during RL framework training to guarantee the training stability and success. For RL framework optimization, only the actor will be updated by an RL training algorithm, namely vanilla policy gradient (VPG), through multiplying the reward with the probability of action leading to that reward. The training objective optimization of VPG is defined as Eq.~\eqref{eq:rltrain}, and the parameter $\phi$ of the policy network $f_\phi(\cdot)$ will be updated.
\begin{align}
\label{eq:rltrain}
&\nabla_{\phi_{\mathcal{B}},\phi_{\mathcal{N}},\phi_{\mathcal{E}}} J = \sum_{t=1}^{T}\left[r(s_t,a_t)\times \nabla_{\phi_{\mathcal{B}},\phi_{\mathcal{N}},\phi_{\mathcal{E}}}(\mathcal{N}(\alpha_i|\mu_i,\Sigma_i) + Ber(\beta_i|p_i))\right] \nonumber \\
& \text{where}\ \alpha_i \sim \mathcal{N}(\cdot|\mu^i,\Sigma^i), \beta_i \sim Ber(\cdot|p_i)
\end{align}

\subsection{Dual-level Training Paradigm}
We develop a dual-level training paradigm for the RL framework training. The meta-training phase aims to expose the RL agent to diverse adversarial attack scenarios so that it can it can gain a general knowledge from learning a wide range of tasks of varying levels of attack, dataset difficulty, input teacher architecture, and computational budget. In the second phase, the agent performs fine-tuning on the target task by leveraging
the meta-trained model as the starting point to achieve quick and
effective adaptation. 

\begin{itemize}[nosep,leftmargin=4mm]
    \item {\em Meta RL training}: In each RL time step as shown in Figure~\ref{fig:RL_train}, first we randomly sample a different task $task=(data,adv,teacher)$ from three initialized data, adversarial attack and architecture buffers. Then, we calculate the Lipschitz coefficient and inference cost encodings (\texttt{LIPS}, \texttt{CT}) and teacher network stage encoding $\texttt{ST}_i, i\in[1,N_{stage}]$. We concatenate these embeddings and pass to the state encoder and get the state $s_t$. Through the policy network $f_\phi(\cdot)$, we get the action $a_t$ for different stages and then conducting RL compression. For the newly RL compressed sub-network, we evaluate its reward using Eq.~\eqref{eq:reward} and train the RL using Eq.~\eqref{eq:rltrain} with VPG. Then, the sub-network will be added into the architecture buffer and move to the next time step by sampling next input task setting. When the maximum time step $T$ is reached, we start a new RL iteration by re-initializing all the buffers until maximum RL iteration $M$ is reached.
    
    \item {\em Downstream RL fine-tuning}: After meta RL training, we let the RL agent quickly adapt to the target domain by conducting a few iterations of fine-tuning. We conduct similar operations as in meta RL training and the only exception is that the target task setting could be repeatedly provided as fine-tuning task input instead of randomly sampling. We set $T=1, M=\Tilde{M}$ for an one time-step, limited iteration RL fine-tuning.
    
\end{itemize}
The psueduo code of the detailed training process is summarized by Algorithms~\ref{alg:n2n} \& ~\ref{alg:downtrain} in the Appendix.

\vspace{-2mm}\subsection{Theoretical Analysis}
   In this section, we theoretically demonstrate how the proposed RL-guided compressed student sub-network trained on adversarial samples results in a better adversarial accuracy compared to the dense network trained using standard adversarial training. To prove this, we leverage the idea of the dense mixture accumulation concept and extend the recently developed theoretical framework that justifies the effectiveness of the adversarial training to improve the robustness performance \cite{allenzhu2022feature}. For this, first, we define the problem setup that involves the key concepts used in our theoretical analysis along with some assumptions to make the proof easier. Next, we present a lemma, demonstrating how the proposed RL-compressed student sub-network leads to the tighter gradient update bound compared to one without compression. Finally, based on the lemma, we present the main theorem that shows how the proposed RL-compressed student sub-network achieves better robust accuracy on adversarial training by ensuring the more reduction in the dense mixture components compared to the the one without compression. For brevity, we denote RL-compressed student sub-network trained using adversarial samples as $S_{RL-C}^A$ and the one without compression as $S_U^A$. 

\vspace{2mm}\noindent {\bf Problem setup.}
Let us assume ${\bf x}_i\in \mathcal{R}^D$ indicates the $i^{th}$ data sample that is used to train the student sub-network obtained using our proposed \texttt{RC-NAS} framework. Also, let us assume that the data sample ${\bf x}_i\in \mathcal{R}^D$ is generated from the sparse coding mechanism with ${\bf x} = {\bf M}{\bf z}+\hat{\epsilon}$, where ${\bf M} \in \mathcal{R}^{D\times D}$ is a sparse matrix whose basis functions is learned by student sub-network. Further ${\bf z}\in \mathcal{R}^D$ is the sparse hidden vector with sparsity defined by parameter $k$. $\hat{\epsilon}\in \mathcal{R}^D$ is the Gaussian noise with zero mean and standard deviation of $\sigma_x$. Also, for the sake of simplicity let us assume that the final student sub-network obtained using \texttt{RC-NAS} is a simple, two-layer symmetric neural network with ReLU activation. Then with $\boldsymbol{\Theta}_{i, t}$ being the  hidden weight for the $i^{th}$ neuron at time $t$, the student sub-network can be represented as:
\begin{align}
   \nonumber & f_t({\Theta}; {\bf x}, \rho) \\
= \; & \sum_{i=1}^N\left(\texttt{ReLU}[\langle\Theta_{i, t},{\bf x}\rangle+\rho_i-b_{t}]-\texttt{ReLU}[-\langle\Theta_{i, t},{\bf x}\rangle+\rho_i-b_{t}]\right)
\end{align}
where $b_{t}$ is the bias at $t^{th}$ at training step $t$, and $\rho_i\sim\mathcal{N}(0, \sigma_p^2)$ is the smoothing of the original ReLU. Then at any clean training $t$, the weight learned by $i^{th}$ neuron can be decomposed as the following
\begin{equation}
    \Theta_{i, t}\approx g_{i, t}+v_{i, t}
\end{equation}
where $g_{i, t} = \mathcal{O}(1) {\bf M}_j$ indicates pure features' contribution to produce the desired output and $v_{i, t} = \sum_{j\prime\neq j}\left[\mathcal{O}\left(\frac{k}{D}\right)\Theta_{j}^{\prime}{\bf M}_{j}^{\prime}\right]$ indicates the dense mixture learned during the clean training in the direction of ${\bf M}_j^\prime$ that are responsible to generate inaccurate response during the adversarial attack. It should be noted that we have a big portion of the robust (pure) features along with some small dense mixtures during the training process in a given neuron. Also, $j\in \mathcal{N}_j$ where $ \mathcal{N}_j$ indicates the subset on which we have a highly correlated pure features. Based on this problem setup we present the following lemma for the gradient update.
\begin{lemma}
    \label{le:gradient_update}
    Let T be the total iterations for the clean training and $T^\prime$ be the additional iterations for the adversarial training. Considering $\delta$ be the l2-perturbation applied on input sample for the adversarial training with a radius of the perturbation defined as $||\delta||_2\leq \tau$. Then gradient movement bound for $S_{RL-C}^A$  is lower than that of the $S_U^A$. Specifically let $\Delta L_t^{RL-C}$ be the gradient movement in  $S_{RL-C}^A$ and $\Delta L_t^{U}$ be the gradient movement in the $S_U^A$, then the following inequality holds at any iteration $t$
    \begin{equation}
        \Delta L_t^{RL-C}\leq \Delta L_t^{U}
    \end{equation}
    
\end{lemma}
\noindent{\bf Remark.} It should be noted that in the case of the $S_{RL-C}^A$, we reduce the width as well as the depth of the given network. This is the same as zeroing out unnecessary edges in our dictionary matrix ${\bf M}$. However, zeroing out the pure (robust) features in dictionary ${\bf M}$ will lead to a smaller reward. Intuitively,  we can infer that to maximize the reward, the RL-agent is forced to compress less important (mostly the dense mixture) components in the given network. Based on this, we have the following theorem.
\begin{theorem}
\label{th: final_dense_mixture}
    Let $v^{T+T^\prime}_{i, RL-C}$ be the final dense mixture component for the $i^{th}$ neuron of $S_{RL-C}^A$   and $v^{T+T^\prime}_{i, U}$ be the dense mixture for the $i^{th}$ neuron of $S_U^A$. Then, based on the gradient update Lemma~\ref{le:gradient_update} with high probability we have the following:
    \begin{equation}
        \max_{i\in N}||v_{i, RL-C}^{T+T^\prime}||_2\leq \max_{i\in N}||v_{i, U}^{T+T^\prime}||_2
    \end{equation}
\end{theorem}
\noindent{\bf Remark.} This theorem indicates that our proposed \texttt{RC-NAS} based sparsification mechanism on a teacher network has the potential to further lower the dense mixture components compared to without sparsification. This is because, through the sparsification, our RL-agent strives to find the sparse student sub-network that can potentially improve the adversarial accuracy by zeroing out the many non-robust entries in the dictionary ${\bf M}$. Please refer to the Appendix for the proof.

\vspace{-2mm}\section{Experiments}

\vspace{2mm}\noindent{\bf Datasets.}
We have three datasets as our test beds, CIFAR-10, CIFAR-100 and Tiny-ImageNet. CIFAR-10 and CIFAR-100 are widely used benchmark datasets in computer vision for image classification tasks. CIFAR-10 consists of 60,000 32x32 color images in 10 classes, with 6,000 images per class, while CIFAR-100 has the same number of images but in 100 classes. In Tiny-ImageNet, there are 100,000 images divided up into 200 classes. Every image in the dataset is downsized to a 64×64 colored image. For every class, there are 500 training images, 50 validating images, and 50 test images.

\vspace{2mm}\noindent{\bf Adversarial attacks.}
Projected Gradient Descent (PGD)~\cite{madry2017towards} is an iterative FGSM method by iteratively conducting FGSM until the image is misclassified or a certain number of iterations is reached.  In our setting, we try $PGD^{20}$ attack methods with an attack radius $\hat{\epsilon}=8/255$ and with a maximum number of iterations $20$. Also, we investigate a traditional Carlini \& Wagner ($CW^{40}$) attack~\cite{carlini2017towards}, which utilizes two separate losses: An adversarial loss to make the generated image actually adversarial, i.e., is capable of fooling image classifiers, and an image distance loss to constrain the quality of the adversarial examples so as not to make the perturbation too obvious to the naked eye. Auto-attack~\cite{croce2020reliable} is a prevailing comprehensive attack method which is a parameter-free, computationally affordable, and user-independent ensemble of existing attacks.

\begin{table*}[!tp]
 \caption{Baseline Comparison on CIFAR datasets} 
 \vspace{-2mm}
\centering
\label{tab: baseline_cifar}
\resizebox{.92\textwidth}{!}{
\begin{tabular}{ccccccccccc}
\toprule
 \multirow{2}{*}{Model} & \multirow{2}{*}{\#P(M)} & \multirow{2}{*}{\#F(G)} & \multicolumn{4}{c}{\bf CIFAR-10} & \multicolumn{4}{c}{\bf CIFAR-100}\\
 \cmidrule{4-11}
& & & Clean & $PGD^{20}$ & $CW^{40}$ & AutoAttack & Clean & $PGD^{20}$ & $CW^{40}$ &  AutoAttack\\
\midrule
WRN-28-10 & 36.5 & 5.20 & 84.62$\pm$0.06 & 55.90$\pm$0.21 & 53.15$\pm$0.33 & 51.66$\pm$0.29 & 56.30$\pm$0.28 & 29.91$\pm$0.40 & 26.22$\pm$0.23 & 25.26$\pm$0.06\\ 
RobNet-large-v2 & 33.3  & 5.10 & 84.57$\pm$0.16 & 52.79$\pm$0.08 & 48.94$\pm$0.04 & 47.48$\pm$0.04 & 55.27$\pm$0.02 & 29.23$\pm$0.15 & 24.63$\pm$0.11 & 23.69$\pm$0.19 \\
AdvRush (7@96) & 32.6 & 4.97 & 84.95$\pm$0.12  & 56.99$\pm$0.08 & 53.27$\pm$0.03 & 52.90$\pm$0.11 & 56.40$\pm$0.09 & 30.40$\pm$0.21 & 26.16$\pm$0.03 & 25.27$\pm$0.02 \\
RACL (7@104)   & 32.5  & 4.93 & 83.91$\pm$0.13 & 55.98$\pm$0.15 & 53.22$\pm$0.08 & 51.37$\pm$0.11 & 56.09$\pm$0.08 & 30.38$\pm$0.03 & 26.65$\pm$0.02 & 25.65$\pm$0.10 \\
RobustResNet-A1  & 19.2  & 5.11 & 85.46$\pm$0.25 & 58.74$\pm$0.12 & 55.72$\pm$0.04 & 54.42$\pm$0.08 & 59.34$\pm$0.09 & 32.70$\pm$0.14 & 27.76$\pm$0.09 & 26.75$\pm$0.14\\
\rowcolor{gray!10} \textbf{RC-NAS} & {\bf 18.8} & \textbf{4.98} & \textbf{86.32$\pm$0.08} & \textbf{60.48$\pm$0.12} & \textbf{58.34$\pm$0.25} & \textbf{57.66$\pm$0.24} & \textbf{62.33$\pm$0.19} & \textbf{34.9$\pm$0.15} & \textbf{29.95$\pm$0.27} & \textbf{29.35$\pm$0.25} \\
\midrule
WRN-34-12 & 66.5 & 9.60 & 84.93$\pm$0.24 & 56.01$\pm$0.28 & 53.53$\pm$0.15 & 51.97$\pm$0.09 & 56.08$\pm$0.41 & 29.87$\pm$0.23 & 26.51$\pm$0.11 & 25.47$\pm$0.10\\ 
WRN-34-R & 68.1 & 19.1 & 85.80$\pm$0.08 & 57.35$\pm$0.09 & 54.77$\pm$0.10 & 53.23$\pm$0.07 & 58.78$\pm$0.11 & 31.17$\pm$0.08 & 27.33$\pm$0.11 & 26.31$\pm$0.03\\
AdvRush (10@96)  & 67.5  & 18.33 & 85.33$\pm$0.13 & 57.08$\pm$0.12 & 54.53$\pm$0.14 & 52.67$\pm$0.15 & 57.14$\pm$0.13 & 30.45$\pm$0.15 & 26.54$\pm$0.15 & 26.22$\pm$0.16\\
RACL (10@104)   & 67.9  & 18.75 & 84.82$\pm$0.18 & 56.38$\pm$0.13 & 53.89$\pm$0.16 & 52.23$\pm$0.16 & 56.78$\pm$0.12 & 30.22$\pm$0.15 &  26.35$\pm$0.17 & 25.79$\pm$0.19\\
RobustResNet-A2  & 39.0  & 10.8 & 85.80$\pm$0.22 & 59.72$\pm$0.15 & 56.74$\pm$0.18 & 55.49$\pm$0.17 & 59.38$\pm$0.15 & 33.0$\pm$0.17 & 28.71$\pm$0.19 & 27.68$\pm$0.21 \\
\rowcolor{gray!10}\textbf{RC-NAS} & \textbf{37.4} & \textbf{9.67} & \textbf{86.84$\pm$0.18} & \textbf{61.08$\pm$0.35} &  \textbf{60.45$\pm$0.24} & \textbf{58.68$\pm$0.15} & \textbf{63.15$\pm$0.22} & \textbf{36.96$\pm$0.25} & \textbf{30.55$\pm$0.36} & \textbf{30.79$\pm$0.33}\\
\midrule
WRN-46-14 & 128 & 18.6 & 85.22$\pm$0.15 & 56.37$\pm$0.18 & 54.19$\pm$0.11 & 52.63$\pm$0.18 & 56.78$\pm$0.47 & 30.03$\pm$0.07 & 27.27$\pm$0.05 & 26.28$\pm$0.03\\
AdvRush (16@100)  & 131 & 23.39 & 86.38$\pm$0.05 & 57.05$\pm$0.12 & 55.08$\pm$0.21 & 54.15$\pm$0.17 & 57.95$\pm$0.28 & 31.25$\pm$0.14 & 28.39$\pm$0.12 & 28.24$\pm$0.13\\
RACL (16@108)   & 132 & 24.12 & 85.45$\pm$0.08 & 56.58$\pm$0.15 & 54.68$\pm$0.24 & 53.29$\pm$0.13 & 57.13$\pm$0.27 & 30.78$\pm$0.17 & 27.85$\pm$0.15 & 27.54$\pm$0.18\\
RobustResNet-A3  & 75.9  & 19.9 & 86.79$\pm$0.09 & 60.10$\pm$0.14 & 57.29$\pm$0.25 & 55.84$\pm$0.15 & 60.16$\pm$0.22 & 33.59$\pm$0.19 & 29.58$\pm$0.12 & 28.48$\pm$0.19 \\
\rowcolor{gray!10}\textbf{RC-NAS} & \textbf{68.5} & \textbf{18.4} & \textbf{88.46$\pm$0.15} & \textbf{62.15$\pm$0.11} & \textbf{60.88$\pm$0.09} & \textbf{59.21$\pm$0.21} & \textbf{64.75$\pm$0.18} & \textbf{37.13$\pm$0.24} & \textbf{31.79$\pm$0.25} & \textbf{31.75$\pm$0.16} \\
\midrule
WRN-70-16 & 267 & 38.8 & 85.51$\pm$0.24 & 56.78$\pm$0.16  & 54.52$\pm$0.16  & 52.80$\pm$0.14  & 56.93$\pm$0.61 & 29.76$\pm$0.17  & 27.20$\pm$0.16 & 26.12$\pm$0.24\\
AdvRush (22@100)  & 266 & 41.75 & 86.11$\pm$0.12 & 56.12$\pm$0.15 & 54.17$\pm$0.09 & 53.38$\pm$0.20 & 56.13$\pm$0.19 & 30.08$\pm$0.16 & 27.16$\pm$0.18 & 27.04$\pm$0.19\\
RACL (22@110)   & 264 & 40.96 & 84.88$\pm$0.24 & 56.17$\pm$0.18 & 54.49$\pm$0.26 & 52.77$\pm$0.14 & 56.79$\pm$0.19 & 30.05$\pm$0.23 & 27.13$\pm$0.14 & 26.88$\pm$0.22\\
RobustResNet-A4  & 147  & 39.4 & 87.10$\pm$0.15 & 60.26$\pm$0.13 & 57.9$\pm$0.18 & 56.29$\pm$0.12 & 61.66$\pm$0.64 & 34.25$\pm$0.19 & 30.04$\pm$0.18 & 29.00$\pm$0.28 \\
\rowcolor{gray!10}\textbf{RC-NAS} & \textbf{129} & \textbf{35.8} & \textbf{89.22$\pm$0.12} & \textbf{62.58$\pm$0.15} & \textbf{61.30$\pm$0.22} & \textbf{59.98$\pm$0.16} & \textbf{65.47$\pm$0.52} & \textbf{37.74$\pm$0.23} & \textbf{31.96$\pm$0.22} & \textbf{32.02$\pm$0.31} \\
\bottomrule
\end{tabular}}

\end{table*}

\vspace{2mm}\noindent{\bf Experimental settings.}
Given four initial teacher networks (WRN-28-10, WRN-34-12, WRN-46-14, WRN-70-16) corresponding to different computation budgets (5G,10G,20G,40G), we apply a range of diverse adversarial attacks targeting on the trained teacher network on the clean training set with the evaluation set and test set data. In RL training and downstream RL finetuning, we leverage the training set and adversarially attacked evaluation set only. Once we get the RL compressed sub-network architectures for the target task, we train it with standard adversarial training on clean training set and test its adversarial performance using classification accuracy on the adversarially attacked test set generated above from the same task setting. For fair comparison, all baselines (architectures found by other research works given same computation budgets) will be trained with standard adversarial training on the clean training set from the same task setting and evaluated their performance on the same test split in the target task setting. The standard adversarial training (AT) method is TRADES~\cite{zhang2019theoretically} with auto-attack as the adversarial attack choice.

\vspace{2mm}\noindent{\bf Baselines.}
We compare our proposed \texttt{RC-NAS} with other latest neural architecture search works for adversarial robustness, such as RobNet-large-v2~\cite{guo2020meets}, AdvRush~\cite{mok2021advrush}, RACL~\cite{dong2020adversarially}, WRN-34-R~\cite{huang2021exploring} and RobustResNet A1-A4~\cite{huang2022revisiting} corresponding to computation budgets 5G, 10G, 20G, 40G, respectively.  We also include the performance of the teacher network in the beginning of each computation budget category for reference.
For fair comparison, we align the network capacity of AdvRush and RACL to different computation budgets by adjusting the number of repetitions of the normal cell N and the input channels of the first normal cell C, denoted as (N@C). The additional details of those baselines are described in Appendix~\ref{app:exp}.

\vspace{-2mm}\subsection{Results and Comparison}
For target task settings, we use CIFAR-10, CIFAR-100 and Tiny-ImageNet as dataset choices, and train all baseline models under four computation budgets using standard adversarial training on the same clean training set and test their adversarial performance on the clean test set or test set with three different adversarial attack categories ($PGD^{20}$, $CW^{40}$, AutoAttack). For the compressed sub-networks selected by the RL agents which are fine-tuned given different target tasks, we train them with standard adversarial training and report its their respective test performance on the corresponding task setting, i.e. CIFAR-10, CIFAR-100 test set with no attack or three different kinds of adversarial attacks, same as other baselines. We summarize the comparison results of CIFAR-10 and CIFAR-100 in Table~\ref{tab: baseline_cifar}, and results of Tiny-ImageNet in Table~\ref{tab: baselinetiny}. From Table~\ref{tab: baseline_cifar} and ~\ref{tab: baselinetiny}, our model consistently achieves better classification accuracy under different computation budgets and adversarial attacks on the test sets, either for CIFAR-10, CIFAR-100 or Tiny-ImageNet, compared to all the other baselines of similar budgets. The RL decided sub-network is not only superior in robust accuracy against adversarial attacks, but also smaller regarding its parameter size (See column $\#P(M)$) and achieves faster inference speed, denoted by GFLOPs in column $\#F(G)$.

\noindent{\bf Remark.} We would like to clarify that we included PGD, which can be regarded as an iterative FGSM. Since it is more severe than FGSM, we did not report FGSM result separately. Instead, we provide a more detailed analysis that varies the PGD attack upper boundary rate from 4/255 to 16/255 and the results are summarized in the table~\ref{tab:pgd}. As can be seen, RC-NAS maintains a clear advantage other strong baselines such as RobustResNet-A1, A2 in 5G/10G computational budgets under different perturbation rates.

\begin{table*}[!ht]
    \centering
    \caption{\label{tab:pgd} Robustness against different PGD attack levels under 5G and 10G budget comparing to RobustResNet-A1,A2 respectively.}
    \begin{tabular}{llllll}
    \toprule
        \textbf{Model} & \textbf{Budget} & \textbf{$\PGD^{20} (\epsilon=4/255)$} & \textbf{$\PGD^{20} (\epsilon=8/255)$} & \textbf{$\PGD^{20} (\epsilon=12/255)$} & \textbf{$\PGD^{20} (\epsilon=16/255)$} \\ \midrule
        \rowcolor{gray!10}RC-NAS & WRN-28-10 (5G) & 61.05$\pm$0.08 & 60.48$\pm$0.12 & 60.32$\pm$0.27 & 60.15$\pm$0.25 \\
        RobustResNet-A1 & 5G & 59.15$\pm$0.13 & 58.74$\pm$0.12 & 58.52$\pm$0.24 & 58.37$\pm$0.26 \\ 
        \rowcolor{gray!10}RC-NAS & WRN-34-12 (10G) & 61.45$\pm$0.29 & 61.08$\pm$0.35 & 60.95$\pm$0.36 & 60.74$\pm$0.39 \\
        RobustResNet-A2 & 10G & 59.79$\pm$0.14 & 59.72$\pm$0.15 & 60.78$\pm$0.25 & 60.39$\pm$0.26 \\
        \bottomrule
    \end{tabular}
\end{table*}

\vspace{-2mm}\subsection{SoTA Adversarial Attacks.} As for black-box attacks, we already included auto-attack~\cite{croce2020reliable}. By investigating latest adversarial attack methods, we conducted experiments by adding advanced boundary attack~\cite{Shen5555}, neuron-based attack~\cite{zhang2022improving} and latest adaptive auto-attack methods ($A^3$)~\cite{yao2021automated}, 
($A3$)~\cite{liu2022practical}. Specifically, we fine-tune our meta-trained RL agent under above SoTA adversarial attacks on CIFAR-10 within WRN-28-10 (5G budget). The comparison result with RobustResNet-A1 is shown in Table~\ref{tab:attacks}. It can be seen that RC-NAS remains robust under these SoTA attacks, and achieves clear advantage comparing to the strongest baseline RobustResNet-A1, thanks to the achieved versatile adversarial robustness.

\begin{table*}[!ht]
    \centering
    \caption{\label{tab:attacks} Robustness against SoTA adversarial attack methods with comparison to RobustResNet-A1.}
    \resizebox{0.75\linewidth}{!}{\begin{tabular}{lllllll}
    \toprule
        \textbf{Model} & \textbf{Budget} & \textbf{	Auto-attack} & \textbf{Boundary-attack}  & \textbf{Neuron-attack }~\cite{zhang2022improving} & \textbf{	$A^3$~\cite{yao2021automated}} & \textbf{$A3$~\cite{liu2022practical}} \\ 
        \midrule
        RC-NAS & WRN-28-10 (5G) & 57.66$\pm$0.24 & 59.66$\pm$0.35 & 60.12$\pm$0.24 & 57.56$\pm$0.28 & 59.98$\pm$0.26 \\ 
        \midrule
        RobustResNet-A1 & WRN-28-10 (5G) & 54.42$\pm$0.08 & 56.15$\pm$0.12 & 56.66$\pm$0.08 &53.56$\pm$0.14 & 56.20$\pm$0.09 \\ \bottomrule
    \end{tabular}}
\end{table*}

\vspace{-2mm}\subsection{RL training and fine-tuning costs.}

Thanks to the unique dual-level training paradigm, the model can learn the key characteristics from different attack scenarios through meta-training, which takes around 45 hours. And by leveraging this general knowledge, the RL agent can perform quick adaptive N2N compression given a specific attack setting in the testing phase, which only takes 10 iterations and 3-9 hours to converge, given different architectures/datasets. For other baselines, to discover suitable sub-networks, they need to randomly sample a large number of architectures and evaluate each one. For example, RobustResNet samples 1,000 subnetworks, easily taking more than 75 hours. More importantly, as shown in our paper, when the learning environment changes, the heuristically derived rules usually lead to suboptimal performance. This implies that a large number of architectures need to be re-sampled and evaluated to meet the new requirements, which is much more expensive than RC-NAS's adaptive training strategy. The fine-tuning time of our models under different data sets, teacher networks with various budgets are detailed in Table~\ref{tab:time}.

\begin{table}[!ht]
    \centering
    \caption{\label{tab:time} Fine-tuning time of our models under different data sets, teacher network with various budgets.}
    \resizebox{0.75\linewidth}{!}{\begin{tabular}{lllll}
    \toprule
        Data  &  Model &  Budget & 	Fine-tuning Time\\ 
        \midrule
        cifar10 & WRN-28-10 & 5 & 3h  \\
        cifar100 & WRN-28-10 & 5 & 5h  \\
        tinyimagenet & WRN-28-10 & 5 & 8h  \\ 
        \midrule
        cifar10 & WRN-34-12 & 10 & 3.5h  \\
        cifar100 & WRN-34-12 & 10 & 5.4h  \\
        tinyimagenet & WRN-34-12 & 10 & 8.2h  \\ \midrule
        cifar10 & WRN-46-14 & 20 & 3.5h  \\
        cifar100 & WRN-46-14 & 20 & 5.6h  \\ 
        tinyimagenet & WRN-46-14 & 20 & 8.8h  \\ \midrule
        cifar10 & WRN-70-16 & 40 & 3.8h  \\ 
        cifar100 & WRN-70-16 & 40 & 6.1h  \\
        tinyimagenet & WRN-70-16 & 40 & 9.5h  \\ \bottomrule
    \end{tabular}}
\end{table}

\begin{table}[tp]
 \caption{Baseline Comparison on Tiny-ImageNet} 
 \vspace{-2mm}
\centering
\label{tab: baselinetiny}
\resizebox{1\linewidth}{!}{
\begin{tabular}{ccccccc}
\toprule
 \multirow{2}{*}{Model} & \multirow{2}{*}{\#P(M)} & \multirow{2}{*}{\#F(G)} & \multicolumn{4}{c}{\bf Tiny-ImageNet}\\
 \cmidrule{4-7}
& & & Clean & $PGD^{20}$ & $CW^{40}$ & AutoAttack\\
\midrule
WRN-46-14 & 128 & 19.8 & 41.23$\pm$0.05 & 20.52$\pm$0.09 & 25.78$\pm$0.12 & 16.28$\pm$0.11 \\
AdvRush (16@100)  & 134 & 22.4 & 41.77$\pm$0.06 & 20.85$\pm$0.12 & 26.19$\pm$0.14 & 17.05$\pm$0.20  \\
RACL (16@108)   & 133  & 22.6 & 41.35$\pm$0.07 & 20.68$\pm$0.08 & 25.93$\pm$0.09 & 16.58$\pm$0.17 \\
RobustResNet-A3  & 75.9  & 20.1 & 47.33$\pm$0.12 & 21.50$\pm$0.14 & 28.92$\pm$0.15 & 25.84$\pm$0.18 \\
\rowcolor{gray!10}\textbf{RC-NAS} & \textbf{72.5} & \textbf{18.4} & \textbf{48.28$\pm$0.17} & \textbf{22.79$\pm$0.12} & \textbf{30.14$\pm$0.18} & \textbf{28.55$\pm$0.14} \\
\midrule
WRN-70-16 & 267 & 38.7 & 42.09$\pm$0.08 & 20.68$\pm$0.07  & 26.02$\pm$0.05  & 16.75$\pm$0.04 \\
AdvRush (22@100)  & 266 & 41.94 & 42.32$\pm$0.09 & 20.82$\pm$0.15 & 26.74$\pm$0.10 & 17.14$\pm$0.18 \\
RACL (22@110)   & 264 & 40.88 & 41.99$\pm$0.08 & 20.74$\pm$0.18 & 26.31$\pm$0.12 & 16.92$\pm$0.15 \\
RobustResNet-A4  & 147  & 39.2 & 49.84$\pm$0.19 & 23.38$\pm$0.15 & 30.56$\pm$0.20 & 27.45$\pm$0.16 \\
\rowcolor{gray!10}\textbf{RC-NAS} & \textbf{145} & \textbf{38.9} & \textbf{50.15$\pm$0.17} & \textbf{23.87$\pm$0.14} & \textbf{31.08$\pm$0.24} & \textbf{29.64$\pm$0.19} \\
\bottomrule
\end{tabular}}

\end{table}

\vspace{-2mm}\subsection{Ablation Study}

\subsubsection{Effectiveness of RL guided exploration}
We investigate the effectiveness of the RL guided architectural exploration by replacing it with other existing techniques, including advanced adversarial training methods (TRADES, SAT, MART) and network pruning methods (Hydra, HARP). In Table~\ref{tab:ablate_rl}, we use Tiny-ImageNet as our test bed and apply 20G, 40G as model's computation budgets. Then, we apply TRADES, SAT, MART to the teacher network using adversarial training and conduct network pruning to the teacher network with score masks (Hydra) or with a holistic aggressive opinion (HARP) to construct baselines. Table~\ref{tab:ablate_rl} clearly shows that the teacher network trained from all non-RL baselines cannot surpass our RL decided sub-network. We further show similar results on cifar datasets in Table~\ref{tab:ablate_rl_cifar} of Appendix~\ref{app:rl_ablate}.

\begin{table}[tp]
    \centering
    \caption{\label{tab:ablate_rl}RL v.s. net pruning and adversarial training baselines}
    \vspace{-2mm}
    \resizebox{1\linewidth}{!}{
    \begin{tabular}{cccccc}
    \toprule
      \multirow{2}{*}{Category} & \multirow{2}{*}{Training Method}  & \multicolumn{4}{c}{\bf Tiny-ImageNet} \\
    \cmidrule{3-6}  
     & & Clean & $PGD^{20}$ & $CW^{40}$ & AutoAttack \\
    \midrule
    \multirow{3}{*}{AT (20G)} & TRADES  & 35.95$\pm$0.11 & 14.12$\pm$0.86 &  16.33$\pm$0.15 & 20.78$\pm$0.32 \\
     &  MART & 32.51$\pm$0.05 & 11.87$\pm$0.12 &  15.13$\pm$0.08 & 17.05$\pm$0.21 \\
     &  SAT & 31.68$\pm$0.06 & 10.99$\pm$0.13 &  14.43$\pm$0.10 & 16.58$\pm$0.25 \\
     \midrule
    \multirow{2}{*}{Network Pruning (20G)} &  Hydra & 35.18$\pm$0.08 & 12.49$\pm$0.54 &  18.59$\pm$0.17 & 17.86$\pm$0.43 \\
      & HARP & 34.77$\pm$0.09 & 11.85$\pm$0.72 &  17.12$\pm$0.84 & 17.08$\pm$0.35 \\
    \midrule
    \multirow{2}{*}{RL (20G)} & R-NAS & 45.12$\pm$1.17 & 17.47$\pm$1.24 & 22.68$\pm$1.29 & 25.94$\pm$1.18 \\
    & \textbf{RC-NAS} & \textbf{48.28$\pm$0.17} & \textbf{22.79$\pm$0.12} & \textbf{30.14$\pm$0.18} & \textbf{28.55$\pm$0.14} \\ 
    \midrule
    \multirow{3}{*}{AT (40G)} & TRADES  & 36.45$\pm$0.10 & 15.07$\pm$0.10 &  17.09$\pm$0.14 & 22.83$\pm$0.34 \\
     &  MART & 32.96$\pm$0.08 & 12.74$\pm$0.11 &  16.05$\pm$0.12 & 19.11$\pm$0.28 \\
     &  SAT & 31.98$\pm$0.05 & 11.72$\pm$0.12 &  15.18$\pm$0.11 & 18.53$\pm$0.26 \\
     \midrule
    \multirow{2}{*}{Network Pruning (40G)} &  Hydra & 35.92$\pm$0.17 & 13.55$\pm$0.65 &  19.55$\pm$0.18 & 19.94$\pm$0.47 \\
      & HARP & 35.24$\pm$0.18 & 12.93$\pm$0.69 &  18.24$\pm$0.99 & 19.15$\pm$0.42 \\
    \midrule
    \multirow{2}{*}{RL (40G)} & R-NAS & 45.18$\pm$1.16 & 18.49$\pm$1.22 & 23.15$\pm$1.30 & 27.13$\pm$1.20 \\
     & \textbf{RC-NAS} & \textbf{50.15$\pm$0.17} & \textbf{23.87$\pm$0.14} & \textbf{31.08$\pm$0.24} & \textbf{29.64$\pm$0.19} \\
     \bottomrule
    \end{tabular}}  
\end{table}

\subsubsection{Effectiveness of dual-level training paradigm}
\begin{figure}
    \centering
    \begin{subfigure}{0.49\linewidth}
      \centering
      \includegraphics[width=\linewidth]{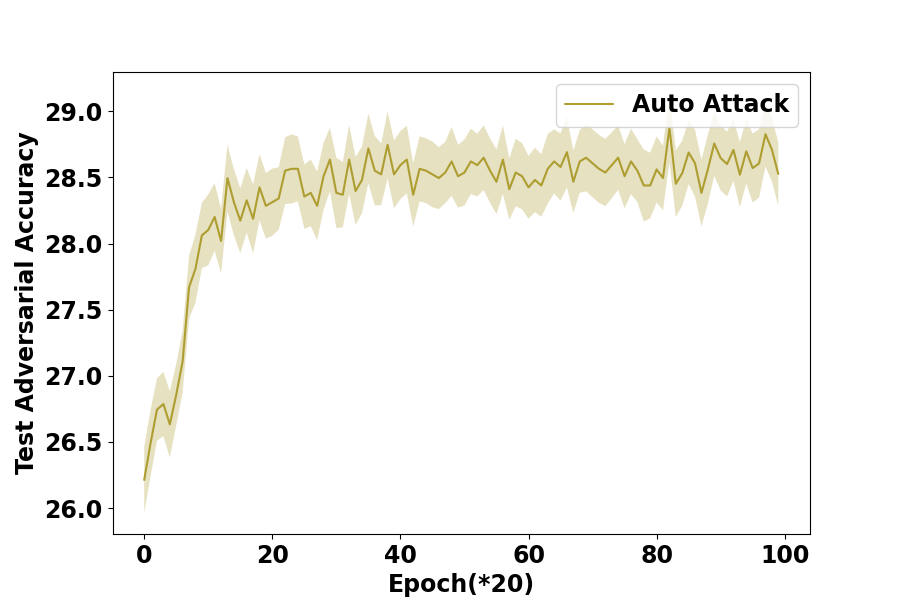}
      \caption{\label{fig:eval_curve_finetune} RL downstream fine-tuned}
    \end{subfigure}
    \begin{subfigure}{0.49\linewidth}
      \centering
      \includegraphics[width=\linewidth]{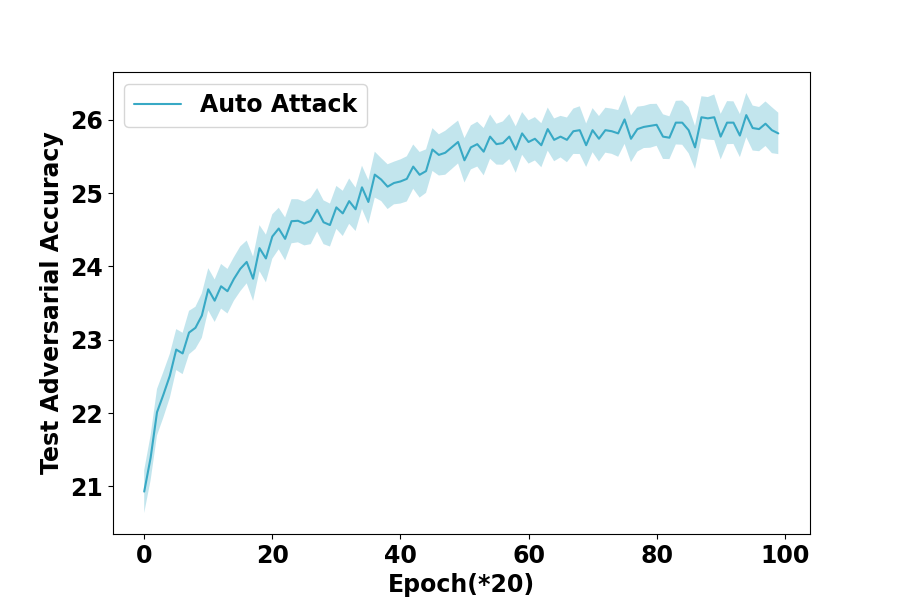}
      \caption{\label{fig:eval_curve_target} RL downstream trained}
    \end{subfigure}
    \vspace{-2mm}
    \caption{\label{fig:eval_curve_tiny}Evaluation curves of the RL (w/ and w/o meta RL training) selected sub-networks on target task setting: WRN-46-14 (20G budget) on auto attacked Tiny-ImageNet.}
    \vspace{-2mm}
\end{figure}

Given the effectiveness of RL mechanism, we further investigate whether the novelly designed dual-level training paradigm helps improve the test performance, which includes a meta RL training phase to optimize under different adversarial tasks and a downstream RL fine-tuning phase to let the meta-trained RL agent quickly adapt to the target task setting. We name the RL agent trained directly on the target task setting (without meta training phase) as R-NAS, and the agent trained with the dual-level training paradigm as RC-NAS. Table~\ref{tab:ablate_rl} shows that on a large dataset Tiny-ImageNet, under same computation budgets, RC-NAS consistently improves over R-NAS, as well as enjoys a lower variance because meta training gives RC-NAS a good weight initialization that can be smoothly adapted into any downstream tasks, without suffering the instability resulting from potentially biased training under a repeated target setting. Such phenomenon has also been verified by the detailed evaluation curves along their respective downstream fine-tuning or training process, where the RL downstream fine-tuned curve with RL meta training will converge fast at earlier training iterations comparing to the RL downstream trained curve w/o meta RL training, shown as Figure~\ref{fig:eval_curve_tiny}.

\subsubsection{Ablative Study on RL critical design components}
We use RL compressed teacher network WRN-46-14 (20G) on Tiny-ImageNet as an example. Given the target task, we first train and downstream fine-tune the RL agent using different design ablation component settings and get the corresponding RL agent with its compressed sub-network. Then, we test the sub-network's classification accuracy on the test set from the same target task, after standard adversarial training. The result is shown in Table~\ref{tab:ablate_component}. We can clearly see that model's performance on all categories will largely drop after the Lipschitz-guided data difficulty embedding $LIPS$ being removed in state formulation. Also, without $Cost$ which reflects the teacher network's inference costs and capacity, the performance will drop a little around 1\%. Annealing represents the model is using soft annealing penalty when constraints are not satisfied, otherwise just assign -1 as a hard penalty for those actions violating the computation budget. By ablating it, the performance also drops around 2\% for different attack categories.

\begin{table}[tp]
    \centering
    \caption{\label{tab:ablate_component}Effectiveness of key components}
    \resizebox{1\linewidth}{!}{
    \begin{tabular}{ccccccc}
    \toprule
     \multicolumn{3}{c}{Component} & \multicolumn{4}{c}{Tiny-ImageNet}\\
     \midrule
    $LIPS$ & $Cost$ & Annealing & Clean & $PGD^{20}$ & $CW^{40}$ & AutoAttack \\
    \midrule
    \XSolidBrush & \Checkmark & \Checkmark & 44.53$\pm$0.15 & 18.78$\pm$0.09 &  27.56$\pm$0.18 & 23.85$\pm$0.15 \\
    \Checkmark & \XSolidBrush & \Checkmark & 47.45$\pm$0.18 & 21.74$\pm$0.11 &  29.65$\pm$0.20 & 27.92$\pm$0.16 \\
    \Checkmark & \Checkmark & \XSolidBrush & 46.29$\pm$0.18 & 20.49$\pm$0.12 &  28.74$\pm$0.19 & 26.88$\pm$0.15 \\
    \rowcolor{gray!10}\Checkmark & \Checkmark & \Checkmark & {\bf 48.28$\pm$0.17} & {\bf 22.79$\pm$0.12} & {\bf 30.14$\pm$0.18} & {\bf 28.55$\pm$0.14}  \\
    \midrule
    \bottomrule
    \end{tabular}}
    
\end{table}

\subsubsection{Statistical Analysis}

\begin{figure}
    \centering
    \begin{subfigure}{0.49\linewidth}
      \centering
      \includegraphics[width=\linewidth]{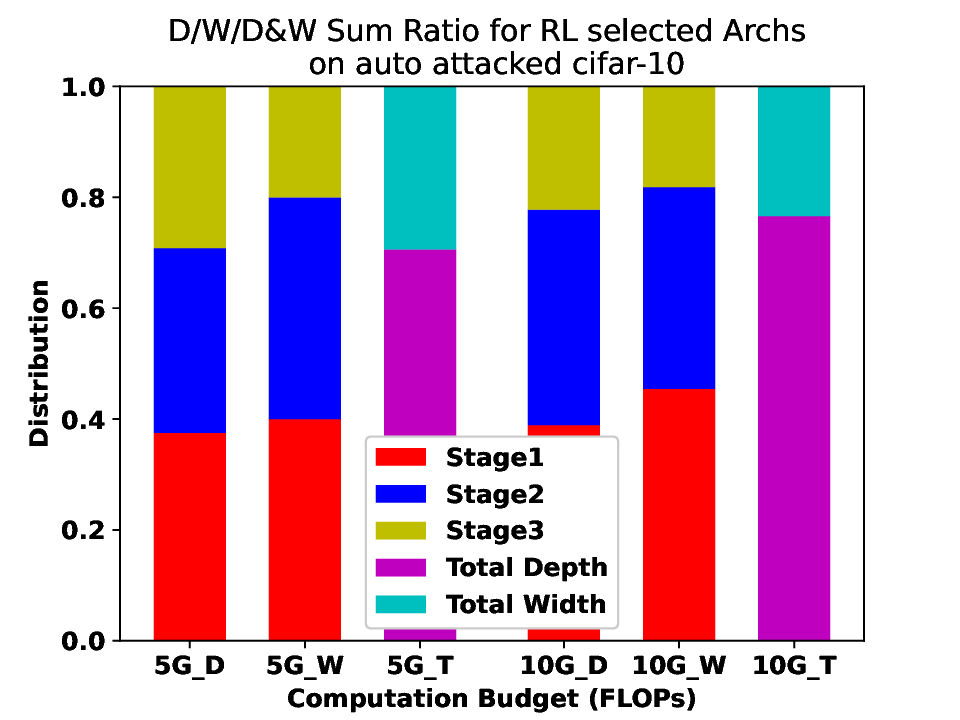}
      \caption{\label{fig:cifar10_pattern} Auto attacked CIFAR-10.}
    \end{subfigure}
    \begin{subfigure}{0.49\linewidth}
      \centering
      \includegraphics[width=\linewidth]{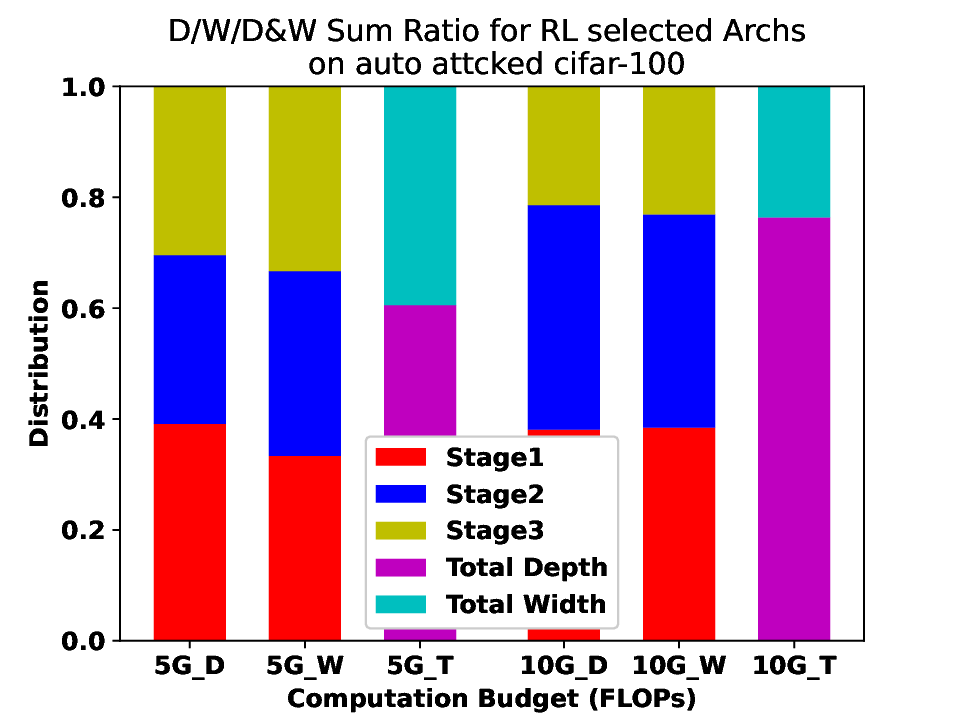}
      \caption{\label{fig:cifar100_pattern} Auto attacked CIFAR-100.}
    \end{subfigure}
    \vspace{-2mm}
    \caption{RL compressed sub-network statistical analysis: depth/width compression ratio across three stages under 5G and 10G budgets (5G-D, 5G-W, 10G-D, 10G-W), and the total depth to total width compression ratio under 5G (5G-T) and 10G (10G-T) budgets on auto attacked CIFAR 10 and 100.}
    \label{fig:pattern}
    \vspace{-2mm}
\end{figure}

Given different teacher networks (WRN-28-10, WRN-34-12) with computation budgets (5G, 10G), we can analyze the RL agent compress decisions on them for auto attacked CIFAR-10 and CIFAR-100 datasets. For WRN-28-10 as teacher network input, it lacks enough generalization ability to auto attack for either CIFAR-10 and CIFAR-100, resulting into a relatively higher width ratio for the last stage comparing to the larger (10G) budget teacher model WRN-34-12, especially for CIFAR-100  which is more complex to classify. For the total width to total depth ratio, 5G model is much higher than 10G model on either CIFAR-10 or CIFAR-100, because the model needs to expand width more to effectively learn the important features for classification given the relatively smaller budget. For the 10G model, it possesses enough parameter learning space and prefers adversarial robustness more than generalization ability, thus tending to reducing the width for higher adversarial robust accuracy. The statistical comparison is shown in Figure~\ref{fig:pattern}.

\section{Conclusion}
In this paper, we propose a \texttt{RC-NAS} framework trained by a novel dual-level training paradigm to achieve reinforced compressive neural architecture architecture search. Specifically, given an input teacher network, a dataset and adversarial attack, our RL agent is able to recognize its difficulty level based upon the capacity of the given teacher network and perform the adaptive stage-level and block-level compression to generate a robust sub-network architecture. Experiments show that our proposed RL sub-network achieves an improved test performance on a wide range of target task's test set across different datasets, adversarial attacks and initial teacher networks. We further investigate the topology of the RL generated sub-network to illustrate its effectiveness in selecting a unique and adversarial robust network architecture given the target task requirements.

\newpage
\bibliographystyle{ACM-Reference-Format}
\bibliography{main}

\newpage
\appendix
\onecolumn

\begin{center}
    {\huge \bf Appendix}
\end{center}

\section{Organization}
We first present all symbols used throughout the paper in Table~\ref{tab:symbols} of Section~\ref{app:symbol}. Next, we illustrate the stage encoder separate pre-training process in Figure~\ref{fig:pretrain} of Section~\ref{app:pre_train} as well as describe the detailed meta RL training and downstream fine-tuning process in Algorithm~\ref{alg:n2n} and ~\ref{alg:downtrain} of Section~\ref{app:alg}. We provide the proof of Lemma~\ref{le:gradient_update} and Theorem~\ref{th: final_dense_mixture} in Appendix~\ref{app:proof}. Then, we provide baseline details and additional experiment results in Appendix~\ref{app:exp}, including the RL mechanism~\ref{app:rl_ablate} and critical design component~\ref{app:component_ablate} ablative study results, as well as more topology comparison~\ref{app:topology} results on CIFAR-10 and CIFAR-100 datasets under different computation budgets. Finally, we discuss the broader impact of our work in Section~\ref{app:impact} and provide a source code link in section~\ref{app:code}.

\section{Summary of Notations}
Table~\ref{tab:symbols} summarizes the major notations used throughout the paper. 
\label{app:symbol}
\begin{table}[!htb]
\caption{\label{tab:symbols} Summary of Notations}
\vspace{-2mm}
\centering
\resizebox{.6\textwidth}{!}{
\begin{tabular}{p{2.5cm}|p{3cm}|p{7cm}}
\Xhline{1.2pt}
\rowcolor{gray!10}\textbf {Symbol Group} &\textbf{Notation} & \textbf{Description} \\
\hline
\cline{1-3}
\multirow{5}{*}{Task Related} & $f_\psi(\cdot)$ & input teacher or generated student network with parameter $\psi$ \\
& $\mathcal{P}$ & adversarial attack method $\mathcal{P}$ \\
& $D, D_{train}, D_{test}, D_{eval}$ & dataset, training, test and evalution set, respectively. \\
& Cost & Inference cost embedding measured by GFLOPs reflecting the teacher network's capacity. \\
& LIPS & Lipschitz-guided embedding representing adversarial attacked dataset difficulty level to the teacher network. \\
\Xhline{1.2pt}
\multirow{15}{*}{RC-NAS} & $SE, DE$ & State encoder and decoder (for pre-training only). \\
& $\theta, \theta_d$ & Parameter space of state encoder and decoder. \\
& $stage_i, N_{stage}$ & $i$th stage encoding and the number of stages in teacher network. \\ 
& $H_i^F, H_i^B$ & Forward and backward LSTM output encodings for $stage_i$ encoding. \\ 
& $f_\phi(\cdot)$ & Multi-head policy network with its parameter space $\phi$. \\
& $\mathcal{N}(\cdot|\mu,\sigma^2),Ber(\cdot|p)$ & Gaussian distribution, Bernoulli distribution. \\
& $\phi_{\mathcal{N}},\phi_{\mathcal{B}},\phi_{\mathcal{E}}$ & Gaussian head, Bernoulli head and their previous shared feature extraction module's respective parameter space. \\
& $\mu_i, \Sigma^2_i$ & Output of the Gaussian head of policy network. \\
& $p_i$ & Output of the Bernoulli head of policy network. \\
& $\hat{\epsilon}, \hat{\sigma}$ & Attack radius (perturbation bound), sigmoid activation function. \\
& $CB, \epsilon$ & Desired computation budget, annealing penalty controlling hyper-parameter when $CB$ is not satisfied. \\
& $\zeta_{RL}$ & Average inference cost of the RL compressed sub-network on the evaluation set. \\
& $\Tilde{A}_{teacher}, \Tilde{A}_{RL}$ & Teacher network and RL compressed sub-network's robust accuracy on the task designated evaluation set. \\
& $C$ & Compression ratio of the RL compressed sub-network compared to its searched teacher network.\\
& $AT$ & Standard Adversarial Training \\
\Xhline{1.2pt}
\multirow{4}{*}{Environment} & $s_t,s_t^i$ & RL state for the whole network, RL state for stage i at time step t. \\
& $a_t,a_t^i$ & RL action for compressing whole network, RL action for compressing stage i at time step t. \\
& $r_t(s_t,a_t)$ & RL reward to evaluation the compression effect for adversarial robustness. \\
& $\mathcal{T}(\cdot|s_t,a_t)$ & RL state transition function, implemented by randomly sampling from three buffers. \\
\Xhline{1.2pt}
\multirow{1}{*}{Theoretical Results} & \\
& $g_{i, t}$ & Pure feature contribution for $i^{th}$ neuron in training iteration $t$.  \\
& $v_{i, t}$ & Dense mixture component for $i^{th}$ neuron in training iteration $t$.  \\
& $\rho_i$ & ReLU smoothing parameter \\
& $b_t$ & Bias at the $t^{th}$ training step \\
& ${\bf M}$ & Sparse matrix from which input is drawn \\
& $\Theta_{i, t}$ & Hidden weight for $i^{th}$ neuron at time t \\
& ${\bf z}$ & Sparse hidden vector with sparsity defined by $k$ \\
& $\hat{\epsilon}$ & Gaussian noise present in the input ${\bf x}$ \\
& $\delta$ & l2-perturbation applied to the input sample \\
& $\tau$ & Radius defining the l2-perturbation \\
& $S_{RL-C}^A$ & Compressed student sub-network obtained using RC-NAS and is trained with adversarial loss \\
& $S_U^A$ & Neural network trained with adversarial loss without RC-NAS \\
& $\Delta L_t^{RL-C}$ & Gradient movement computed in $S_{RL-C}^A$ \\
& $\Delta_t^{U}$ & Gradient movement computed in $S^A_U$ \\
& $v_{i, RL-C}^{T+T^\prime}$ & Final dense mixture component for the $i^{th}$ neuron of $S_{RL-C}^A$ network \\
& $v_{i, U}^{T+T^\prime}$ & Final dense mixture component for $i^{th}$ neuron of $S_U^A$ network \\
& T & Total clean training iterations \\
& $T^\prime$ & Additional Adversarial training performed on the top of clean training \\ 
\Xhline{1.2pt}
\end{tabular}}
\end{table}

\section{State Encoder Pre-training}
\label{app:pre_train}

\begin{figure*}[h!]
    \centering
    \includegraphics[width=0.8\linewidth]{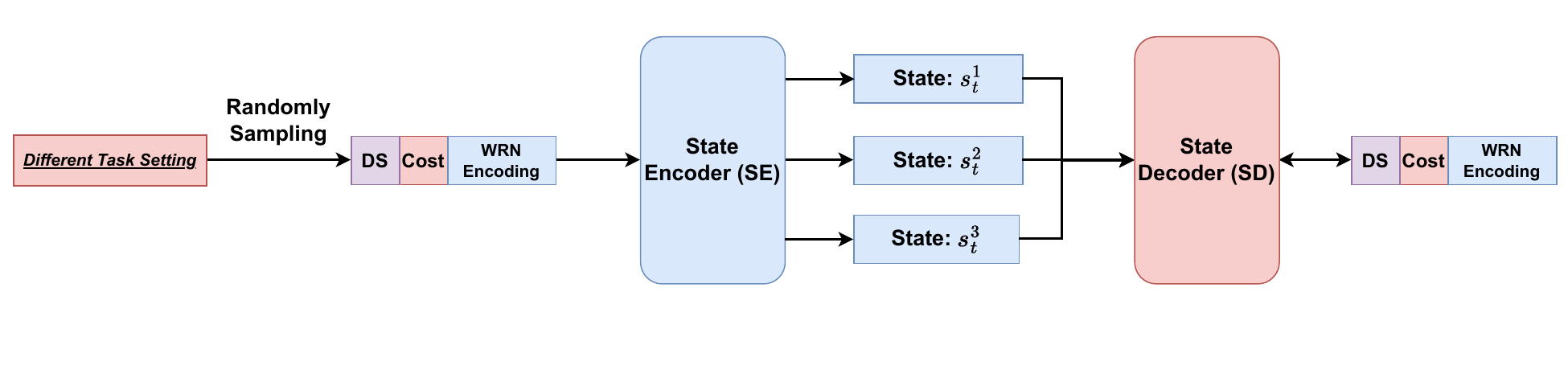}
    \caption{\label{fig:pretrain}State Encoder Pre-Training}
\end{figure*}

Figure~\ref{fig:pretrain} shows the process of the state encoder training. Specifically, we pre-train it with randomly sampled diverse tasks, where the stage encodings $stage_i$, data difficulty embedding $LIPS$ and inference cost embedding $Cost$ are input into the state encoder to generate the state $s^i_t$. Then we decode $s^i_t$ into the same inputs again with an additional decoder using the supervision by the input $\texttt{concat}(\texttt{LIPS},\texttt{ST}_i,\texttt{CT})$ itself. The training target is formulated as Equation~\eqref{eq:pretrain}, where $DE_{\theta_d}(s^i_t)$ means the decoder network.

\begin{align}
\label{eq:pretrain}
\mathcal{L}_\text{SE} = (\texttt{concat}(\texttt{LIPS},\texttt{ST}_i,\texttt{CT}) 
- DE_{\theta_d}(SE_\theta(\texttt{LIPS},\texttt{ST}_i,\texttt{CT})))^2
\end{align}

\section{Meta RL training and downstream RL fine-tuning}
\label{app:alg}
The general RL training process is included in Algorithms~\ref{alg:n2n}. For meta-training, we set 100 iterations for the RL agent to learn the general knowledge. Each iteration includes 5 time steps so 500 architectures will be trained. State and action generation are fast as they only perform forward passes. Reward evaluates current selected sub-network. We train each such subnetwork in 5 epochs and evaluate their accuracy to get the reward. 

\begin{algorithm}[ht]
\caption{\label{alg:n2n} Meta RL Training for Compressive Neural Architecture Searching.}
\begin{algorithmic}

\State Meta RL training:
\Require dataset Buffer $\mathbbm{B}_{data}$; Adversarial Attack Buffer $\mathbbm{B}_{adv}$; Architecture Buffer $\mathbbm{B}_{arch}$; Initialize State Encoder $SE(\cdot|\theta)$, policy Network $f_\phi(\cdot)$, RL total iteration number M, RL total time step T
\For{m = 1 to M} 
\State Initialize architecture buffer $\mathbbm{B}_{arch}$ with a pre-defined teacher network pool
\State Sample input dataset $D$ from buffer $\mathbbm{B}_{data}$
\State Sample adversarial attack method $A$ from buffer $\mathbbm{B}_{adv}$
\State Split $D$ into training $D_{train}$ and evaluation $D_{eval}$, respectively.

\For {t = 1 to T}
\State Sample a teacher network architecture $Net_{Teach}$ from architecture buffer $\mathbbm{B}_{arch}$.
\State Train the teacher network on clean training set $D_{train}$ with standard adversarial training $AT$.
\State Adversarial attack the teacher network on evaluation set $D_{eval}$ with attack choice $A$ to construct $A$-attacked evaluation set $D_{eval}$.
\State Input $D_{eval},A,Net_{Teach}$ into Stage Encoder $SE(\cdot|\theta)$, encode WRN stages $Stage_{i}, i\in[1,N_{stage}]$, data difficulty $LIPS$~\eqref{eq:lipschitz} and inference cost $Cost$ and get the output state $s_t$ using Equation~\eqref{eq:state}.
\State Get the action $a_t\sim f_\phi(s_t)$ which designates newly RL compressed network configuration for each stage with Equation~\eqref{eq:action}. 
\State Generate the RL compressed network $Net_{Stu}$ and store it into the architecture buffer $\mathbbm{B}_{arch}$
\State Train RL compressed network $Net_{Stu}$ with standard adversarial training $AT$ on clean training set $D_{train}$ and evaluation it on adversarial evaluation set $D_{eval}$ with the same attack choice $A$, the evaluated adversarial robust accuracy $\Tilde{A}_{RL}$ is used to form the RL reward $r_t(s_t,a_t)$, with Equation~\eqref{eq:reward}.
\State Pre-train the state encoder on combined training and evaluation set $D_{train},D{eval}$ with Eq.~\eqref{eq:pretrain}, then freeze its weight and optimize the policy network parameters $\phi$ using Eq.~\eqref{eq:rltrain}.
\EndFor 
\EndFor 
\State \textbf{Output}: Meta-Trained RL framework, i.e. RL trained state encoder $SE(\cdot|\theta^{\prime})$ and policy network $f_{\phi^{\prime}}(\cdot)$.

\end{algorithmic}
\end{algorithm}

\begin{algorithm}
\caption{\label{alg:downtrain} Downstream RL fine-tuning and testing for target task setting.}
\begin{algorithmic}
\Require Meta-trained State Encoder $SE(\cdot|\theta)$, policy Network $f_\phi(\cdot)$, RL total iteration number $\Tilde{M}$, RL total time step $\Tilde{T}=1$, target task setting including dataset ($D_{train}$,$D_{test}$,$D_{eval}$), adversarial attack $A$ and initial teacher network $Net_{Teach}$ for compression.

\State Adversarial attack the evaluation set, test set $D_{eval},D_{test}$ with designated task parameters: Attack choice $A$ targeting standard adversarial trained teacher network $Net_{Teach}$ on training set$D_{train}$.
\State Pre-train the state encoder on combined training and evaluation set $D_{train},D{eval}$ with target adversarial attack $A$ and initial network $Net_{Teach}$ using Eq.~\eqref{eq:pretrain}, then freeze its weight for later inference use.

\For{m = 1 to $\Tilde{M}$}
\For {t = 1 to $\Tilde{T}$}
\State Input $D_{eval},A,Net_{Teach}$ into Stage Encoder $SE(\cdot|\theta)$, encode teacher network $Net_{teach}$ stages $Stage_{i}, i\in[1,N_{stage}]$, data difficulty $LIPS$~\eqref{eq:lipschitz} and inference cost $Cost$ to get the output state $s_t$ using Equation~\eqref{eq:state}.
\State Get the action $a_t\sim P_S(s_t|\phi)$ which designates newly RL compressed network configuration for each stage with Equation~\eqref{eq:action}. 
\State Generate the RL compressed network $Net_{Stu}$ and store it into the architecture buffer $\mathbbm{B}_{arch}$
\State Train RL compressed network $Net_{Stu}$ with standard adversarial training on clean training set $D_{train}$ and evaluation it on adversarial evaluation set $D_{eval}$ from the target task setting, the evaluated adversarial robust accuracy $\Tilde{A}_{RL}$ is used to form the RL reward $r_t(s_t,a_t)$, with Equation~\eqref{eq:reward}.
\State Optimize (fine-tuning) the policy network parameters $\phi$ using Eq.~\eqref{eq:rltrain}.
\EndFor 
\EndFor

\State \textbf{Output}: Downstream finetuned RL agent for the target domain task complexity, with trained state encoder $SE(\cdot|\theta^{\prime})$ and policy network $f_{\phi^{\prime}}(\cdot)$.

\State (Optional) For RL testing:
\State Leverage downstream fine-tuned RL agent to sample an RL compressed sub-network from teacher network $Net_{Teach}$, then train it with standard adversarial training on clean training set from the target task, then test it on adversarial attacked test set $D_{test}$ from the same target task setting, for adversarial performance evaluation.
\end{algorithmic}
\end{algorithm}

\section{Theoretical Proof}
\label{app:proof}
In this section, we provide the Proof of Lemma~\ref{le:gradient_update} and Theorem~\ref{th: final_dense_mixture}. Before formally proving them, first we would define the gradient update in our case which would be the essential block to prove our Lemma and Theorem.
\subsection{Gradient Computation}
At every iteration t, the gradient of the loss with respect to the neuron $\Theta_i$ can be represented as :
\begin{equation}
    \Delta_{{\Theta}_{i, t}} = -yl^\prime({\Theta_{i, t}; {\bf x}, y, \rho})\left(\mathbbm{1}_{\langle\Theta_{i, t},{\bf x}\rangle]+\rho_i\geq b_{t}}+\mathbbm{1}_{-\langle\Theta_{i, t},{\bf x}\rangle]+\rho_i\geq b_{t}}\right)\cdot {\bf x}
\end{equation}
For simplicity, considering the task to be binary classification problem i.e., $y\in [-1, 1]$, and loss to be log-loss represented as:
\begin{equation}
    l(\Theta_t; {\bf x}, y, \rho) = \log\left(1+\exp(-yf_t(\Theta; {\bf x}, \rho))\right)
\end{equation}
Further, $ l^\prime({\Theta_{i, t}; {\bf x}, y, \rho})$ is defined as:
\begin{equation}
    l^\prime({\Theta_{i, t}; {\bf x}, y, \rho}) = \frac{\delta}{\delta s} [\log (1+\exp(s)]|_{s=-yf_t(\Theta; {\bf x}, \rho)}
\end{equation}

\subsection{Proof of Lemma~\ref{le:gradient_update}}
Suppose at any training iteration t, $\max_{i\in [N]}||v_i||_2\leq r^\prime$. Also, for the sake of simplicity let $l({\Theta_{t}; {\bf x}, y, \rho}) = l \in [-1, 1]$. Also, consider $\delta\in \mathcal{R}^D$ be the perturbation applied on ${\bf x}$ with $||\delta||_2\leq \tau$. 
Then, l2 norm of average gradient direction bound for $i^{th}; i\in [N]$ neuron in the  network with the adversarial input ${\bf x}+\delta$ can be represented as:
\begin{equation}
    B = ||\mathbbm{E}_{{\bf x}, y, \rho}\left[l\mathbbm{1}_{\langle g_i+v_i, {\bf x}+\delta\rangle+\rho_i\geq b}({\bf x}+\delta)\right]||_2
\end{equation}
It should be noted that for the sake of simplicity, we have removed the subscript $t$ wherever applicable.
For the generic gradient direction shown above, we will first derive the upper bound and then instantiate to the RL-guided compressed sub-student network $S_{RL-C}^A$ and dense network without compression i.e., $S_U^A$. It is worth mentioning that we apply adversarial training on both networks. 
Using the Analysis of F.4 from \cite{allenzhu2022feature}, we can write the following
\begin{equation}
    ||\mathbbm{E}[l\mathbbm{1}_{\langle g_i+v_i, {\bf x}+\delta \rangle+\rho_i\geq b}\delta]||_2\leq \kappa \tau 
    \label{del_bound}
\end{equation}
In the above equation $\kappa = \mathcal{O}(\frac{k}{D}+\frac{(r^\prime)^2}{db^2})$
Now let us try to find the bound for the norm of $\boldsymbol{\beta} = \mathbbm{E}[l\mathbbm{1}_{\langle g_i+v_i, {\bf x}+\delta\rangle+\rho_i\geq b}{\bf x}]$

According to Eq. F.7 \cite{allenzhu2022feature}, inner product of $\boldsymbol{\beta}$ with ${\bf M}_j; j\in [D]$ can be decomposed into the following three terms:
\begin{equation}
    |\langle \boldsymbol{\beta}, {\bf M}_j\rangle||\leq \mathbbm{E}\left[\mathbbm{1}_{\langle g_i, {\bf x} \rangle\geq \frac{b}{10}}+\mathbbm{1}_{\langle v_i, {\bf x}-{\bf M}_j{\bf z}_j\rangle\geq \frac{b}{20}}.|\langle {\bf x}, {\bf M}_j|\right]+\frac{1}{poly(D)}
\end{equation}
The summation of the first term in above equation for all neurons can be represented as (using Eq. F.8 from \cite{allenzhu2022feature})
\begin{equation}
    \sum_{j\in [D]}\left(\mathbbm{E}[\mathbbm{1}_{\langle g_i, {\bf x}\rangle\geq \frac{b}{10}}].|\langle {\bf x}, {\bf M}_j|\right)^2\leq s\mathcal{O}\left(\frac{k}{D^2}\right)
    \label{first_term}
\end{equation}
It should be noted that the scalar s takes different values depending on the nature of the training. In the case of the RC-NAS student sub-network adversarial training, as the model is forced to zero out the weights, $\langle{\bf x}, {\bf M}_j\rangle$ will always be less compared to the dense network trained using adversarial training.  In other words we $u = 1$ in the case of $S_U^A$, and  $0\leq u\leq 1$ in the case of the $S_{RL-C}^A$.   

Similarly for the second term, applying the fact that $\langle{\bf M}_j, \hat{\epsilon}\rangle$ for RC-NAS based student sub-network adversarial training is less than that of the standard adversarial training on the dense network, we have following (according to F.9 from \cite{allenzhu2022feature}):
\begin{equation}
    \sum_{j\in [D]} \left(\mathbbm{E}\left[\mathbbm{1}_{\langle v_i, {\bf x}-{\bf M}_j{\bf z}_j\geq \frac{b}{20}}.|\langle{\bf x}, {\bf M_j}\rangle|\right]\right)^2 \leq w\left(\frac{(r^\prime)^4}{D^2b^4}.\left(\frac{k}{D}+\sigma_x^2\log^2D\right)\right)
\label{second_term}
\end{equation}
Here $w = 1$ for the $S_U^A$ whereas, $0 \leq w\leq 1$ for $S_{RL-C}^A$.

Similarity the bound for the third term with leveraging the fact that $\langle v_i, {\bf M}_j\rangle$ is smaller for the RC-NAS student sub-network compared to the standard adversarial dense network training, we have following (using F.10 in \cite{allenzhu2022feature})
\begin{equation}
    \sum_{j\in [D]}\left([\mathbbm{1}_{\langle v_i, {\bf M}_j\rangle {\bf z}_j\geq \frac{b}{20}}].|\langle{\bf x}, {\bf M}_j\rangle|\right)\leq u\left(\frac{({r}^\prime)^2}{D^2b^2}\right)
    \label{third_term}
\end{equation}
In case of $S_{RL-C}^A$, $0\leq u\leq 1$ whereas, $u = 1$ for $S_U^A$.
Combining Eqs~\ref{first_term}, \ref{second_term}, \ref{third_term} and \ref{del_bound}, we have the following 
\begin{equation}
   || \mathbbm{E}_{{\bf x}, a, \rho}\left[l\mathbbm{1}_{\langle g_i+v_i, {\bf x}, \delta\rangle+\rho_i\geq b}({\bf x+\delta})\right]||_2\leq c \mathcal{O}\left(\left(\frac{k}{D}+\frac{(r^\prime)^2}{Db^2}\right)\tau+\frac{\sqrt{k}}{d}+\frac{(r^\prime)^2}{Db^2}\left(\frac{\sqrt{k}}{\sqrt{D}}+\sigma_xlogD\right)+\frac{r^\prime}{Db}\right)
\end{equation}
In the above equation $c=1$ for $S_U^A$ whereas $0\leq c\leq 1$ for the $S_{RL-C}^A$. Considering left hand side of the above Equation as $\Delta L_t$ with $\Delta L_t^{RL-C}$ being  gradient bound for $S_{RL-C}^A$ and $\Delta L^{U}_t$  being the gradient bound for $S_U^A$,  we can write 
\begin{equation}
        \Delta L_t^{RL-C}\leq \Delta L_t^{U}
        \label{lm1_eq}
    \end{equation}
    This completes the Lemma~\ref{le:gradient_update}. 
\subsection{Proof of Theorem~\ref{th: final_dense_mixture}}
Considering the Equation~\ref{le:gradient_update} is true in Lemma~\ref{le:gradient_update} then the $v_i^{T+T^\prime}$ after $T^\prime$ steps of training can be expressed as 
\begin{equation}
    ||v_i^{T+T^\prime}||_2\leq  ||v_i^{T}||_2+T^\prime\eta c\mathcal{O} \left(\left(\frac{k}{D}+\frac{(r^\prime)^2}{Db^2}\right)\tau+\frac{\sqrt{k}}{D}+\frac{(r^\prime)^2}{Db^2}\left(\frac{\sqrt{k}}{\sqrt{D}}+\sigma_xlogD\right)+\frac{r^\prime}{Db}\right)
\end{equation}
Where $\eta$ is the learning rate used in the SGD to update network. It should be noted that $c=1$ for the standard dense network adversarial training i.e., on a network $S_U^A$ whereas $0\leq c\leq 1$ in the RC-NAS guided sparse adversarial training  on student sub-network i.e., on  $S_{RL-C}^A$. This means to show $\max_{i\in [N]}||v_i^T+T^\prime||\leq r^\prime$, we can choose the following
\begin{equation}
   c\mathcal{O} \left(\left(\frac{k}{D}+\frac{(r^\prime)^2}{Db^2}\right)\tau+\frac{\sqrt{k}}{D}+\frac{(r^\prime)^2}{Db^2}\left(\frac{\sqrt{k}}{\sqrt{D}}+\sigma_xlogD\right)+\frac{r^\prime}{Db}\right)\leq r^\prime 
\end{equation}
Using this we can conclude 
\begin{equation}
    r^\prime_{U}\geq r^\prime_{RL-C}\
\end{equation}
Where $ r^\prime_{U}$ is the upper bound of dense mixture component for $S_U^A$  and $r^\prime_{RL-C}$ be the dense mixture upper bound for $S_{RL-C}^A$.
By using this inequality, we can write 
 \begin{equation}
        \max_{i\in N}||v_{i, RL-C}^{T+T^\prime}||_2\leq \max_{i\in N}||v_{i, U}^{T+T^\prime}||_2
    \end{equation}
    Thie completes the proof for Theorem~\ref{th: final_dense_mixture}.

\section{Additional Experiment}
\label{app:exp}

\subsection{Baseline Description}
Guo et al. ~\cite{guo2020meets} take an architectural perspective and investigate the patterns of network architectures that are resilient to adversarial attacks. To obtain the large number of networks needed for this study, they adopt one-shot neural architecture search, training a large network for once and then finetuning the sub-networks sampled therefrom, the best of the model series searched by this method is called RobNet-large-v2. Dong et al.~\cite{dong2020adversarially} explore the relationship among adversarial robustness, Lipschitz constant, and architecture parameters and show that an appropriate constraint on architecture parameters could reduce the Lipschitz constant to further improve the robustness, namely RACL. Mok et al.~\cite{mok2021advrush} propose AdvRush, a novel adversarial robustness-aware neural architecture search algorithm, based upon a finding that independent of the training method, the intrinsic robustness of a neural network can be represented with the smoothness of its input loss landscape. Through a regularizer that favors a candidate architecture with a smoother input loss landscape, AdvRush successfully discovers an adversarially robust neural architecture. Specifically, we align the network complexity of AdvRush and RACL models by adjusting the number of repetitions of the normal cell N and the input channels of the first normal cell C, denoted as (N@C). Huang et al.~\cite{huang2021exploring} propose WRN-34-R based on three key observations derived via a comprehensive investigation on the impact of network width and depth on the robustness of adversarially trained DNNs. Additionally, in the latest work~\cite{huang2022revisiting}, Huang et al. propose a portfolio of adversarially robust residual networks, dubbed RobustResNets A1-A4, spanning a broad spectrum of model FLOP budgets (i.e., 5G - 40G FLOPs), based on a series of architecture search rules found by a large-scale architecture investigation on CIFAR-10 dataset.

\subsection{AT/NP/Meta Ablative Baseline Comparison on cifar datasets}
\label{app:rl_ablate}
We further collect the adversarial training, network pruning and RL based methods test performance comparison results on a wide range of target task settings, including two cifar datasets, four clean/adversarial attack categories and two initial teacher network with 5G and 10G computation budgets. All non-RL based methods follow the same training and test procedures in the same task setting as compared RL method for fair comparison. We find that with RL dual-level training mechanism, the proposed RC-NAS is consistently better than other non-RL based baselines or the baseline without meta RL training, which aligns with the conclusion in Table~\ref{tab:ablate_rl}.

\begin{table}[tp]
    \centering
    \caption{RL v.s. network pruning v.s. adversarial training baselines}
    \label{tab:ablate_rl_cifar}
    \resizebox{1\linewidth}{!}{
    \begin{tabular}{cccccccccc}
    \toprule
      \multirow{2}{*}{Category} & \multirow{2}{*}{Training Method}  & \multicolumn{4}{c}{\bf CIFAR-10} & \multicolumn{4}{c}{\bf CIFAR-100}\\
    \cmidrule{3-10}  
     & & Clean & $PGD^{20}$ & $CW^{40}$ & AutoAttack & Clean & $PGD^{20}$ & $CW^{40}$ & AutoAttack\\
    \midrule
    \multirow{3}{*}{AT (5G)} & TRADES  & 84.62$\pm$0.06 & 55.90$\pm$0.21 & 53.15$\pm$0.33 & 51.66$\pm$0.29 & 60.98$\pm$0.07 & 32.79$\pm$0.09 & 27.28$\pm$0.12 & 24.94$\pm$0.14\\
     &  MART & 81.29$\pm$0.15 & 52.85$\pm$0.40 & 51.36$\pm$0.33 & 48.74$\pm$0.27 & 57.29$\pm$0.23 & 29.12$\pm$0.25 & 27.48$\pm$0.20 & 18.94$\pm$0.22\\
     &  SAT & 80.87$\pm$0.12 & 52.44$\pm$0.36 & 50.97$\pm$0.09 & 48.25$\pm$0.24 & 56.88$\pm$0.21 & 28.76$\pm$0.22 & 26.52$\pm$0.18 & 18.23$\pm$0.19\\
     \midrule
    \multirow{2}{*}{Network Pruning (5G)} &  Hydra & 84.14$\pm$0.07 & 53.79$\pm$0.18 &  58.47$\pm$0.20 & 47.15$\pm$0.05 & 60.79$\pm$0.08 & 31.23$\pm$0.12 & 29.82$\pm$0.10 & 21.68$\pm$0.05\\
      & HARP & 83.12$\pm$0.05 & 52.65$\pm$0.08 &  57.12$\pm$0.06 & 45.98$\pm$0.04 & 59.84$\pm$0.09 & 30.17$\pm$0.09 & 28.95$\pm$0.08 & 20.73$\pm$0.07\\
    \midrule
      \multirow{2}{*}{RL (5G)} & R-NAS & 83.12$\pm$0.23 & 56.74$\pm$0.15 &  55.69$\pm$0.22 & 54.12$\pm$0.23 & 59.19$\pm$0.21 & 33.42$\pm$0.18 & 27.68$\pm$0.16 & 26.14$\pm$0.27 \\
      & \textbf{RC-NAS} & \textbf{86.32$\pm$0.08} & \textbf{60.48$\pm$0.12} & \textbf{58.34$\pm$0.25} & \textbf{57.66$\pm$0.24} & \textbf{62.33$\pm$0.19} & \textbf{34.9$\pm$0.15} & \textbf{29.95$\pm$0.27} & \textbf{29.35$\pm$0.25}\\
    \midrule
    \multirow{3}{*}{AT (10G)} & TRADES  & 84.98$\pm$0.07 & 56.45$\pm$0.20 & 53.79$\pm$0.35 & 52.12$\pm$0.21 & 61.96$\pm$0.15 & 33.47$\pm$0.18 & 27.98$\pm$0.11 & 25.89$\pm$0.16\\
     &  MART & 81.98$\pm$0.16 & 53.38$\pm$0.25 & 51.96$\pm$0.29 & 49.42$\pm$0.18 & 57.89$\pm$0.29 & 29.98$\pm$0.26 & 28.15$\pm$0.19 & 19.75$\pm$0.23\\
     &  SAT & 81.88$\pm$0.16 & 53.12$\pm$0.38 & 51.93$\pm$0.19 & 48.98$\pm$0.22 & 57.79$\pm$0.22 & 29.45$\pm$0.24 & 27.31$\pm$0.20 & 18.99$\pm$0.23\\
     \midrule
    \multirow{2}{*}{Network Pruning (10G)} &  Hydra & 84.98$\pm$0.15 & 54.23$\pm$0.17 &  59.19$\pm$0.23 & 47.74$\pm$0.06 & 61.52$\pm$0.13 & 31.85$\pm$0.11 & 30.59$\pm$0.11 & 22.74$\pm$0.16\\
      & HARP & 83.58$\pm$0.06 & 52.95$\pm$0.10 &  57.74$\pm$0.07 & 46.85$\pm$0.12 & 60.04$\pm$0.05 & 30.54$\pm$0.11 & 29.31$\pm$0.12 & 21.76$\pm$0.15\\
    \midrule
    \multirow{2}{*}{RL (10G)} & R-NAS & 83.12$\pm$0.23 & 61.74$\pm$0.15 &  56.69$\pm$0.22 & 44.12$\pm$0.23 & \textbf{62.68$\pm$0.14} & \textbf{36.05$\pm$0.19} & \textbf{29.13$\pm$0.16} & \textbf{29.85$\pm$0.17} \\
      & \textbf{RC-NAS} &  \textbf{86.84$\pm$0.18} & \textbf{61.08$\pm$0.35} &  \textbf{60.45$\pm$0.24} & \textbf{58.68$\pm$0.15} & \textbf{63.15$\pm$0.22} & \textbf{36.96$\pm$0.25} & \textbf{30.55$\pm$0.36} & \textbf{30.79$\pm$0.33}\\
    \bottomrule
    \end{tabular}}

\end{table}

\begin{table}[tp]
    \centering
    \caption{\label{tab:ablate_component_cifar} RC-NAS with different design component ablations under 10G budget.}   
    \resizebox{\linewidth}{!}{
    \begin{tabular}{ccccccccccc}
    \toprule
     \multicolumn{3}{c}{Component} & \multicolumn{4}{c}{CIFAR-10} & \multicolumn{4}{c}{CIFAR-100}\\
     \midrule
    $LIPS$ & $Cost$ & Annealing & Clean & $PGD^{20}$ & $CW^{40}$ & AutoAttack & Clean & $PGD^{20}$ & $CW^{40}$ & AutoAttack\\
    \midrule
    \XSolidBrush & \Checkmark & \Checkmark & 80.13$\pm$0.14 & 51.28$\pm$0.08 &  55.98$\pm$0.09 & 44.14$\pm$0.12 & 59.25$\pm$0.12 & 33.85$\pm$0.14 & 27.42$\pm$0.18 & 24.17$\pm$0.19 \\
    \Checkmark & \XSolidBrush & \Checkmark & 86.42$\pm$0.17 & 54.59$\pm$0.19 &  60.05$\pm$0.14 & 48.29$\pm$0.18 & 63.21$\pm$0.18 & 35.74$\pm$0.21 & 30.13$\pm$0.24 & 26.14$\pm$0.32 \\
    \Checkmark & \Checkmark & \XSolidBrush & 84.65$\pm$0.22 & 53.12$\pm$0.28 &  58.39$\pm$0.26 & 46.71$\pm$0.14 & 61.32$\pm$0.24 & 34.88$\pm$0.22 & 29.04$\pm$0.28 & 24.95$\pm$0.31 \\
    \rowcolor{gray!10} \Checkmark & \Checkmark & \Checkmark & \textbf{86.84$\pm$0.18} & \textbf{61.08$\pm$0.35} &  \textbf{60.45$\pm$0.24} & \textbf{58.68$\pm$0.15} & \textbf{63.15$\pm$0.22} & \textbf{36.96$\pm$0.25} & \textbf{30.55$\pm$0.36} & \textbf{30.79$\pm$0.33} \\
    \midrule
    \bottomrule
    \end{tabular}}

\end{table}

\begin{figure}
    \centering
    \begin{subfigure}{0.24\linewidth}
      \centering
      \includegraphics[width=\linewidth]{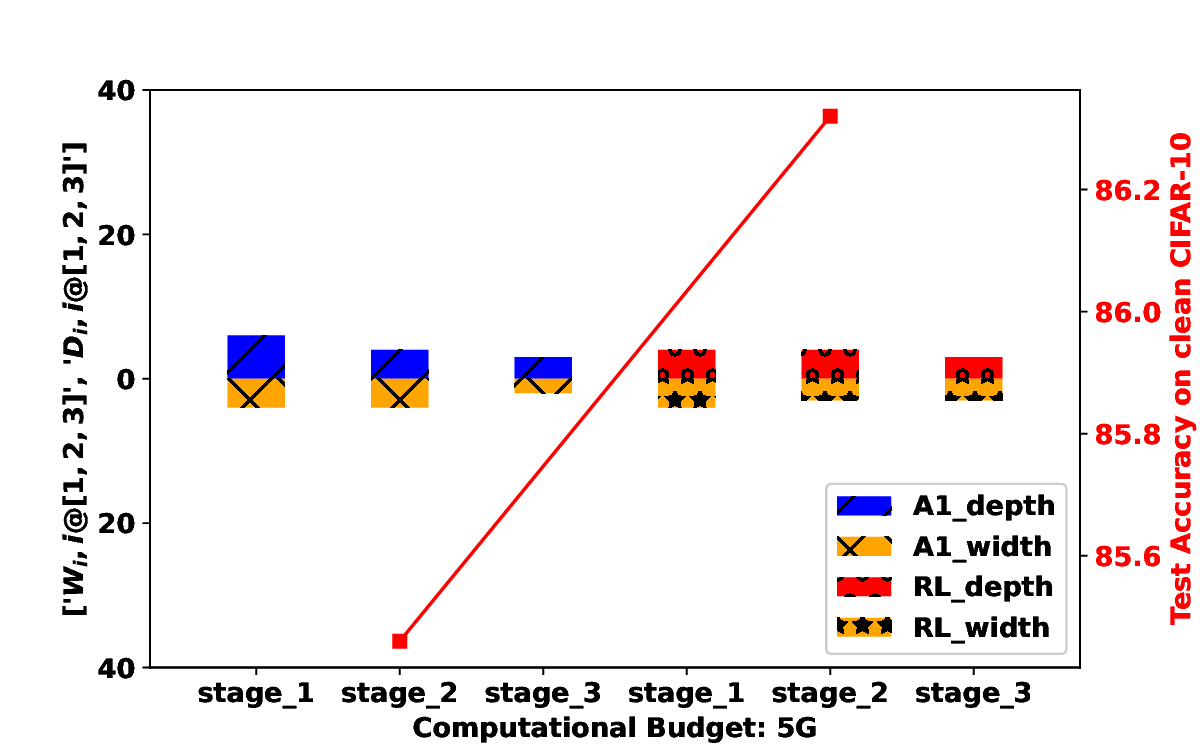}
      \caption{\label{fig:cifar10_5g_clean} 5G on clean }
    \end{subfigure}
    \begin{subfigure}{0.24\linewidth}
      \centering
      \includegraphics[width=\linewidth]{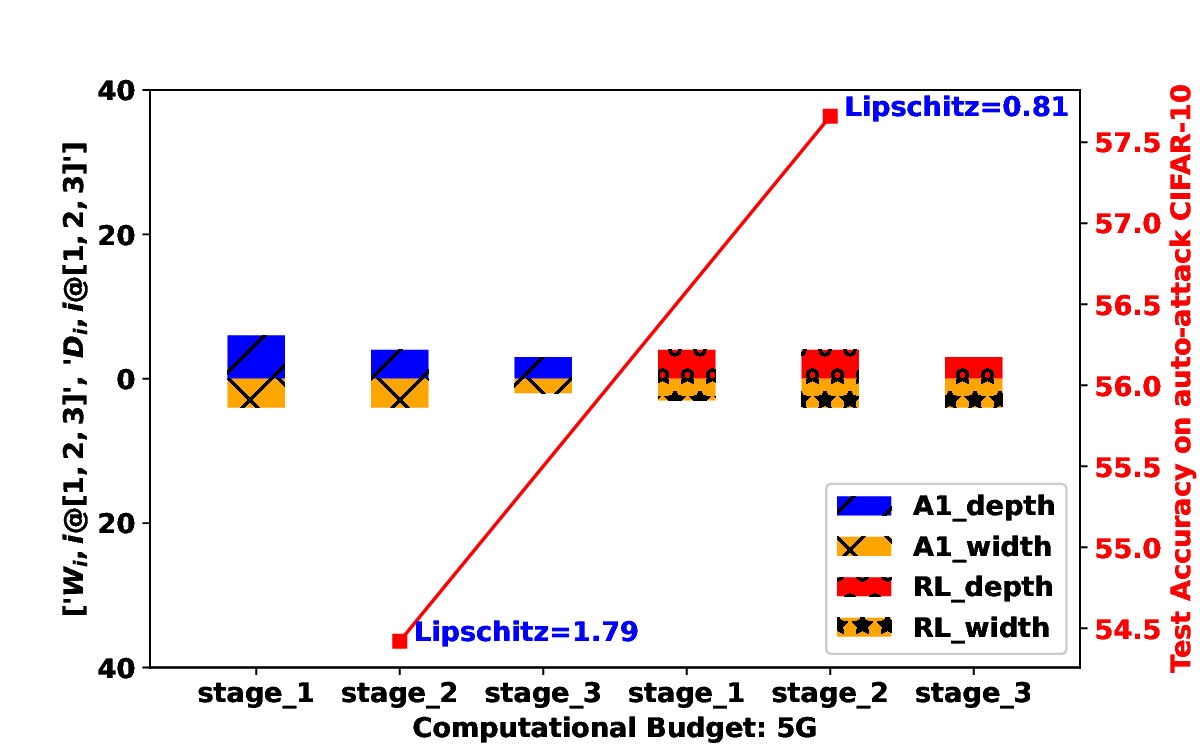}
      \caption{\label{fig:cifar10_5g_auto} 5G on auto-attack }
    \end{subfigure}
    \begin{subfigure}{0.24\linewidth}
      \centering
      \includegraphics[width=\linewidth]{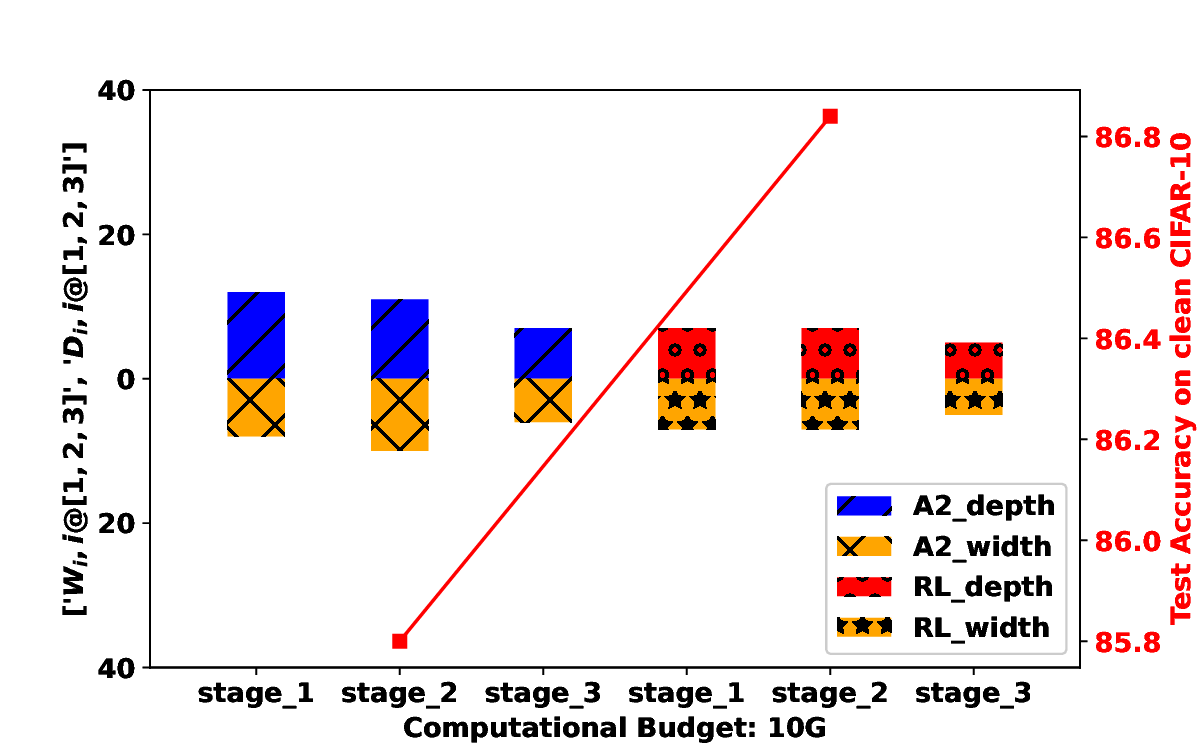}
      \caption{\label{fig:cifar10_10g_clean} 10G on clean }
    \end{subfigure}
    \begin{subfigure}{0.24\linewidth}
      \centering
      \includegraphics[width=\linewidth]{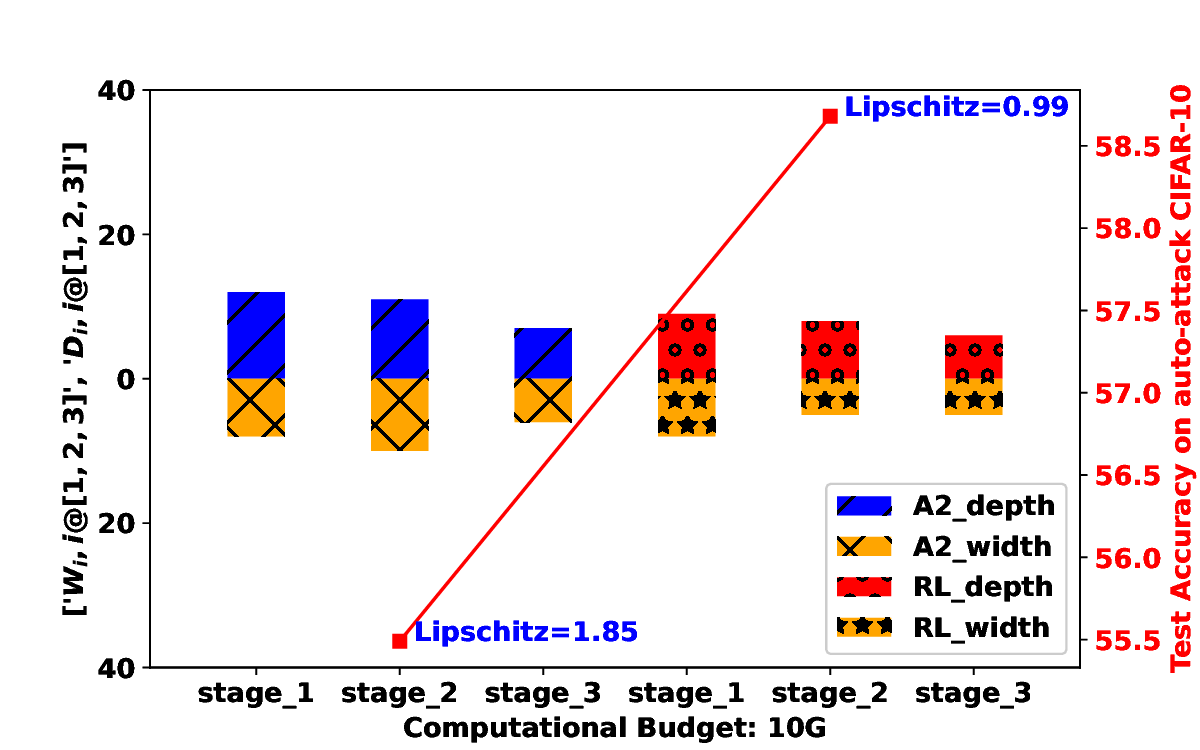}
      \caption{\label{fig:cifar10_10g_auto} 10G on auto-attack }
    \end{subfigure}\\
    \begin{subfigure}{0.24\linewidth}
      \centering
      \includegraphics[width=\linewidth]{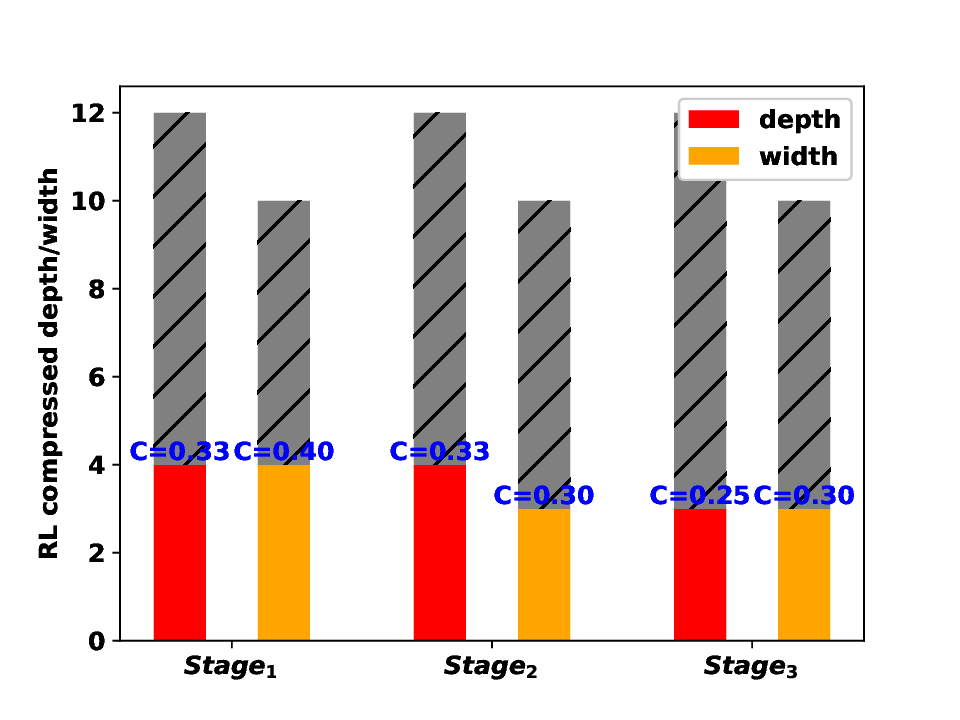}
      \caption{\label{fig:cifar10_5g_clean_rl} WRN-28-10 on clean }
    \end{subfigure}
    \begin{subfigure}{0.24\linewidth}
      \centering
      \includegraphics[width=\linewidth]{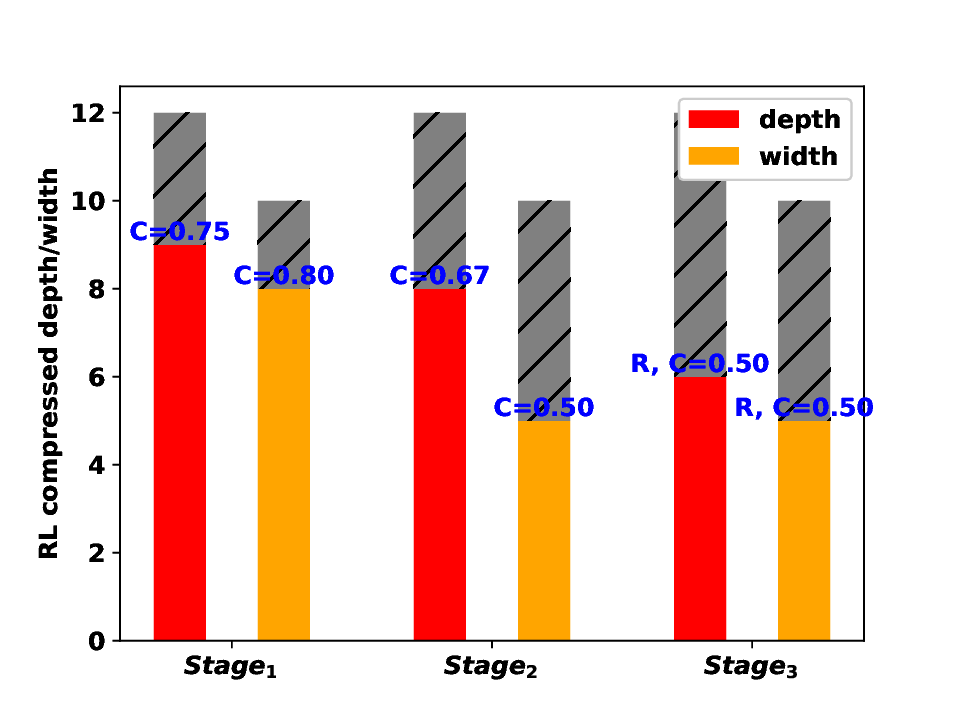}
      \caption{\label{fig:cifar10_5g_auto_rl} WRN-28-10 on auto-attack }
    \end{subfigure}
    \begin{subfigure}{0.24\linewidth}
      \centering
      \includegraphics[width=\linewidth]{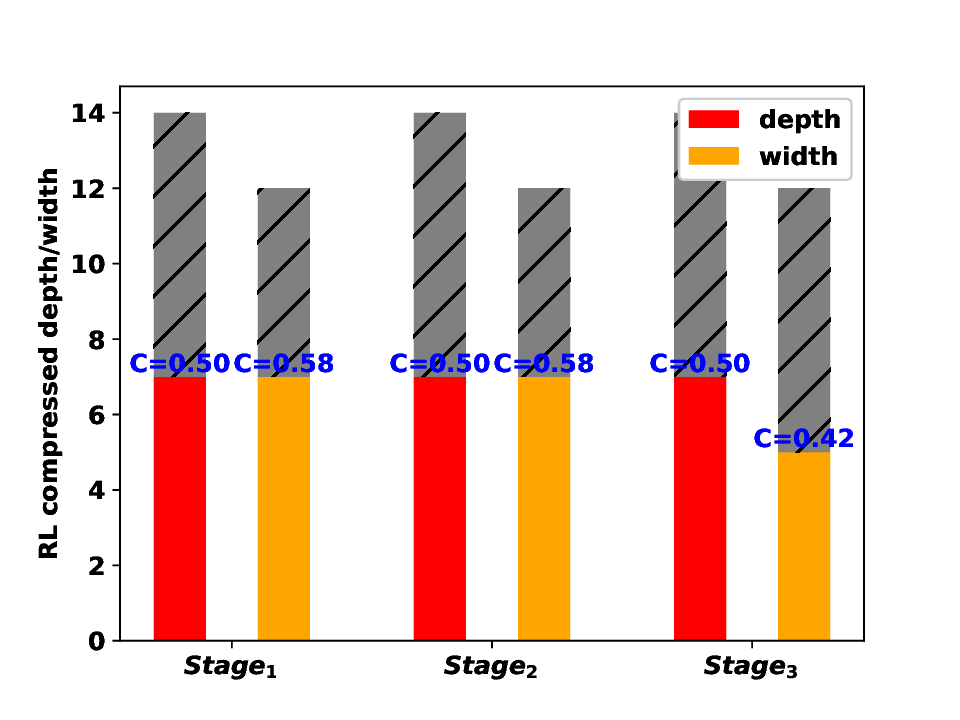}
      \caption{\label{fig:cifar10_10g_clean_rl} WRN-34-12 on clean }
    \end{subfigure}
    \begin{subfigure}{0.24\linewidth}
      \centering
      \includegraphics[width=\linewidth]{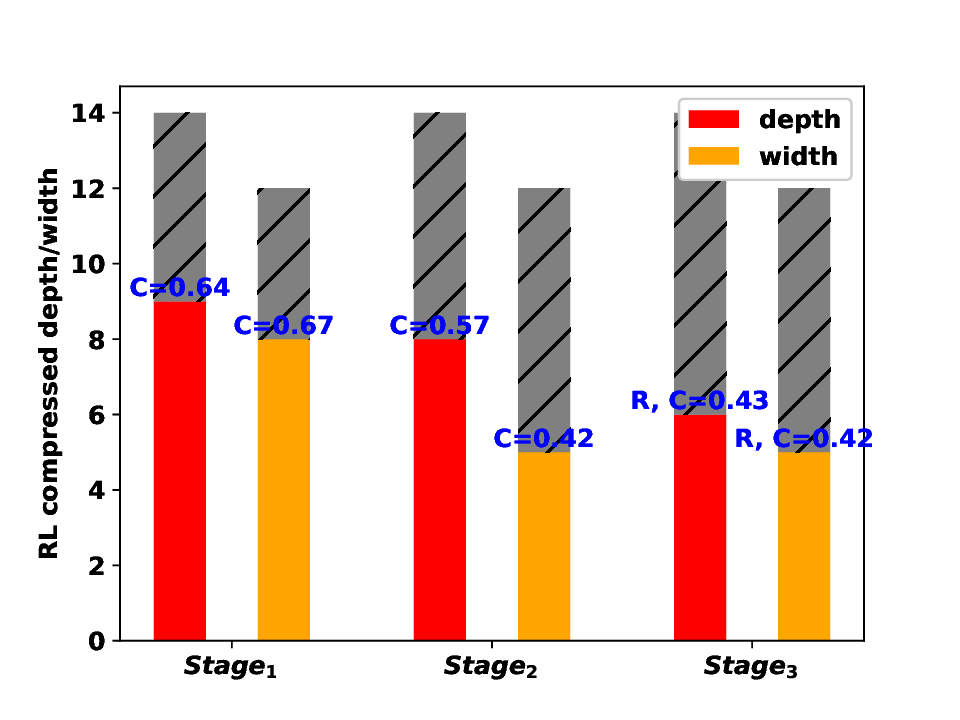}
      \caption{\label{fig:cifar10_10g_auto_rl} WRN-34-12 on auto-attack }
    \end{subfigure}\\
    \begin{subfigure}{0.24\linewidth}
      \centering
      \includegraphics[width=\linewidth]{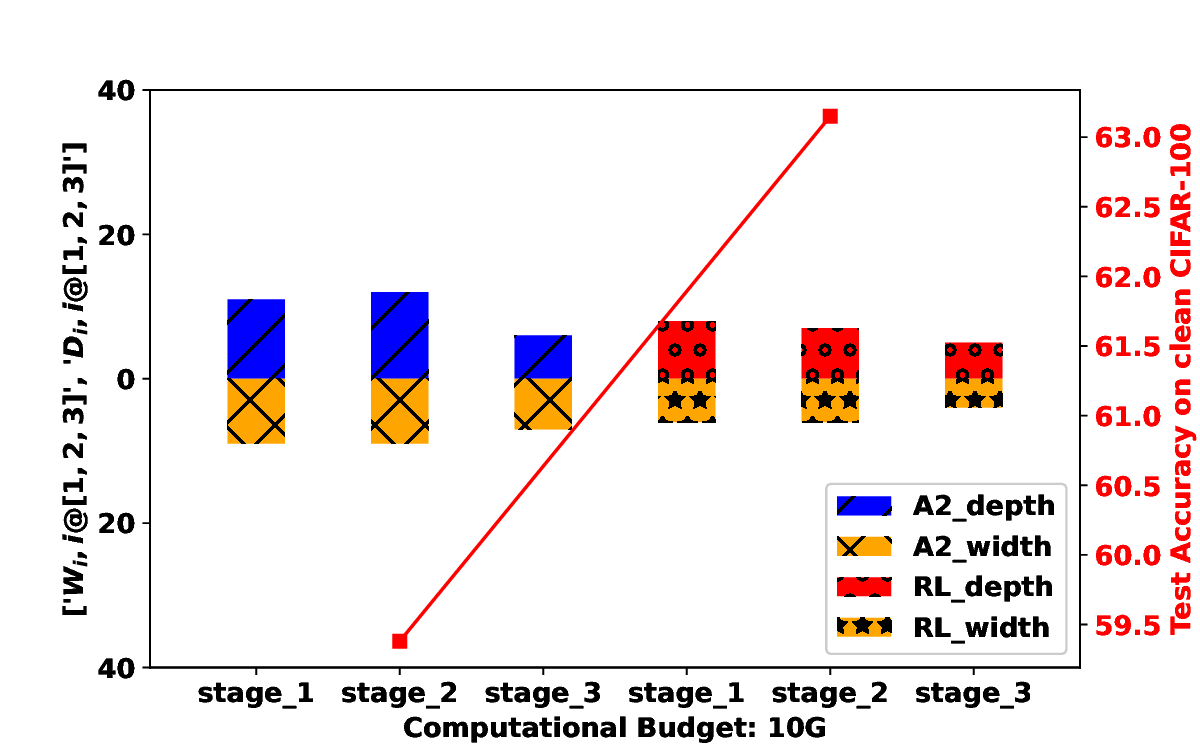}
      \caption{\label{fig:cifar100_10g_clean} 10G on clean }
    \end{subfigure}
    \begin{subfigure}{0.24\linewidth}
      \centering
      \includegraphics[width=\linewidth]{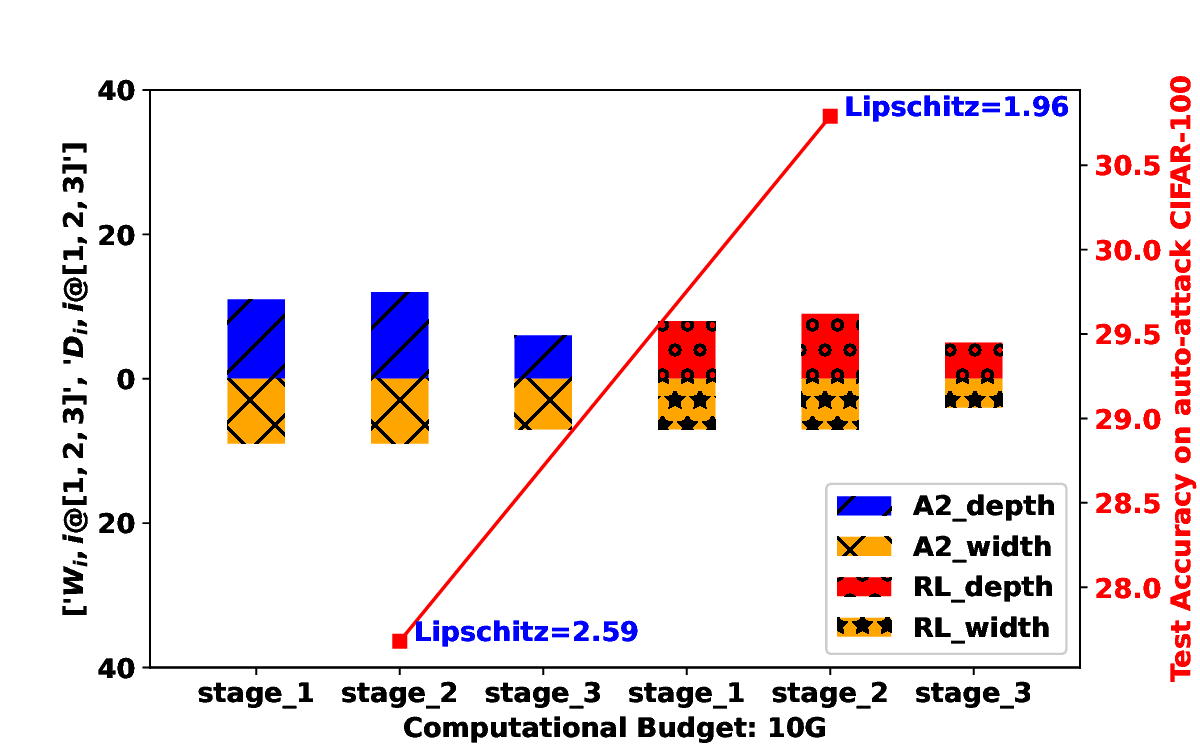}
      \caption{\label{fig:cifar100_10g_auto} 10G on auto-attack }
    \end{subfigure}
    \begin{subfigure}{0.24\linewidth}
      \centering
      \includegraphics[width=\linewidth]{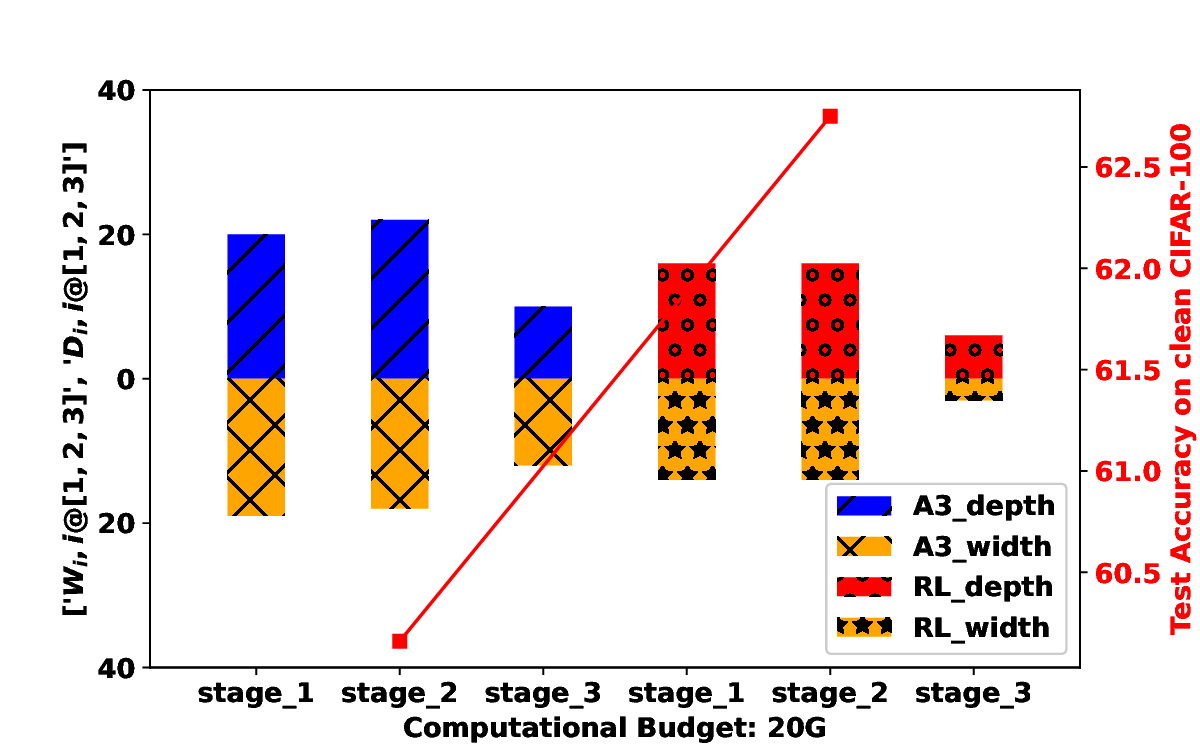}
      \caption{\label{fig:cifar100_20g_clean} 20G on clean }
    \end{subfigure}
    \begin{subfigure}{0.24\linewidth}
      \centering
      \includegraphics[width=\linewidth]{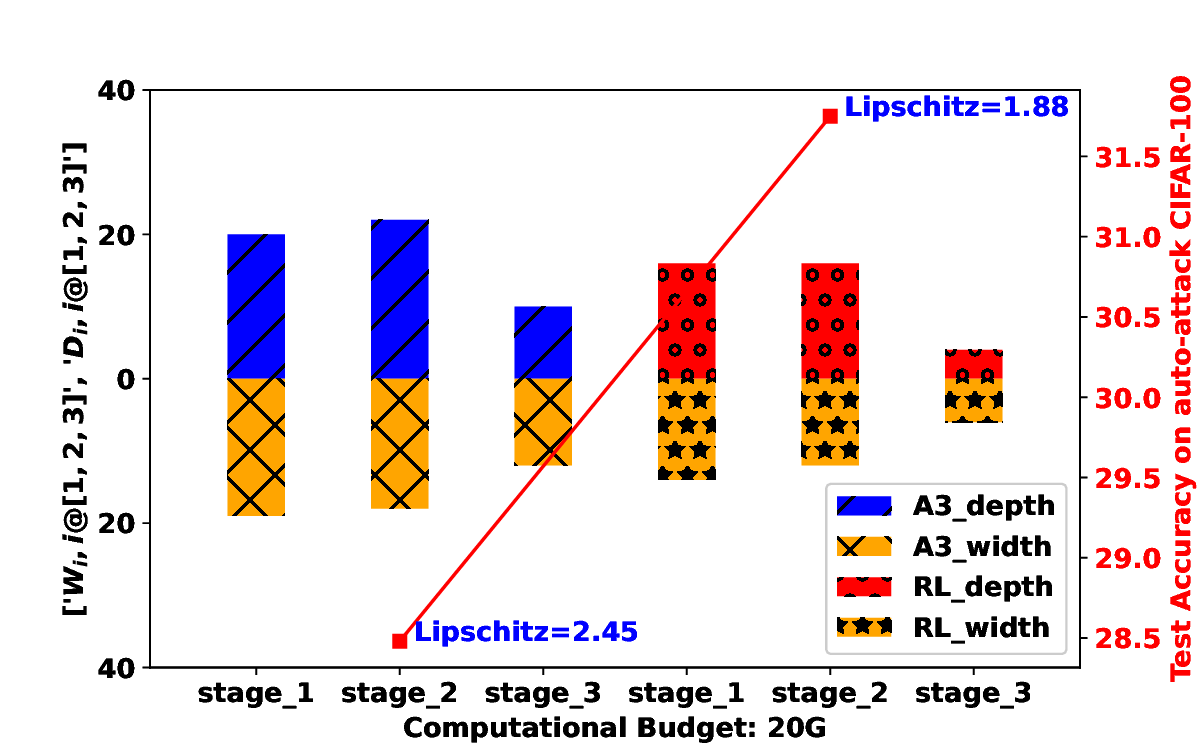}
      \caption{\label{fig:cifar100_20g_auto} 20G on auto-attack }
    \end{subfigure}\\
    \begin{subfigure}{0.24\linewidth}
      \centering
      \includegraphics[width=\linewidth]{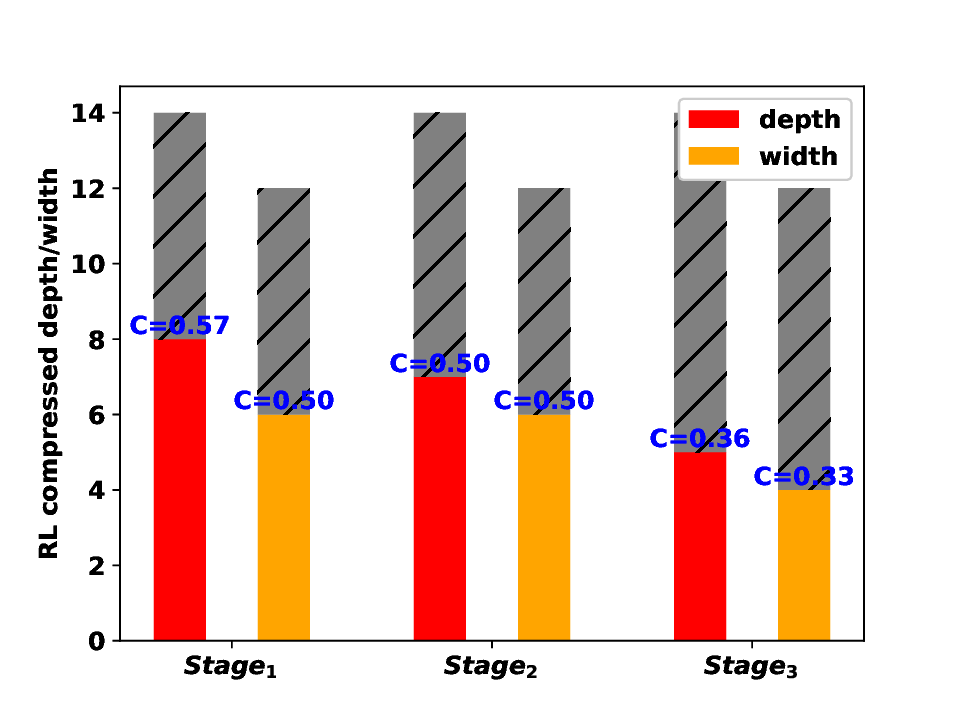}
      \caption{\label{fig:cifar100_10g_clean_rl} WRN-34-12 on clean }
    \end{subfigure}
    \begin{subfigure}{0.24\linewidth}
      \centering
      \includegraphics[width=\linewidth]{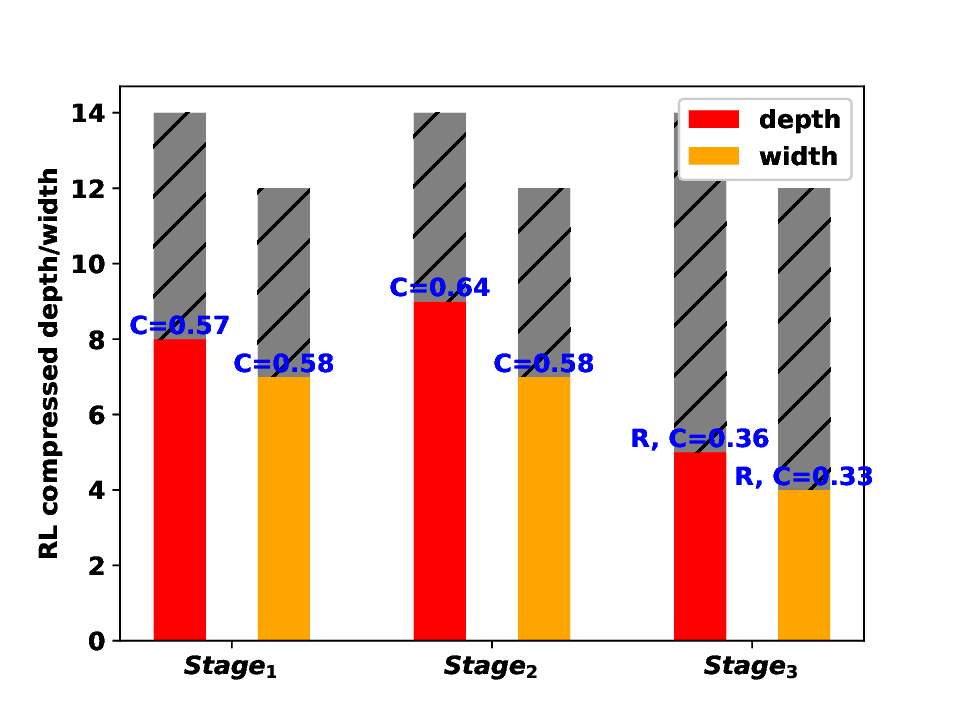}
      \caption{\label{fig:cifar100_10g_auto_rl} WRN-34-12 on auto-attack }
    \end{subfigure}
    \begin{subfigure}{0.24\linewidth}
      \centering
      \includegraphics[width=\linewidth]{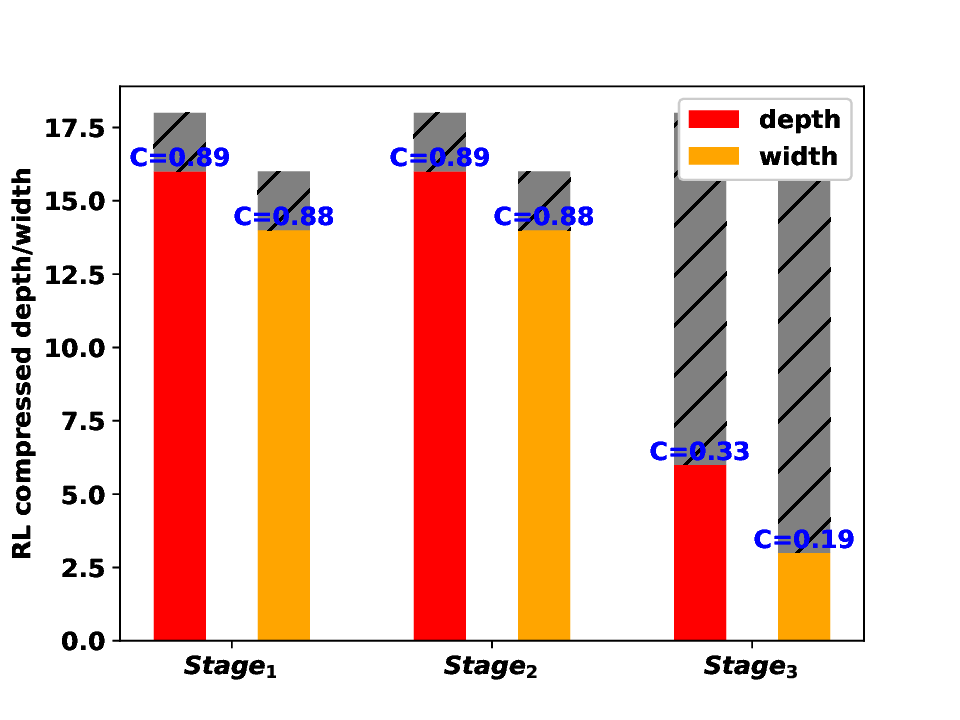}
      \caption{\label{fig:cifar100_20g_clean_rl} WRN-46-14 on clean }
    \end{subfigure}
    \begin{subfigure}{0.24\linewidth}
      \centering
      \includegraphics[width=\linewidth]{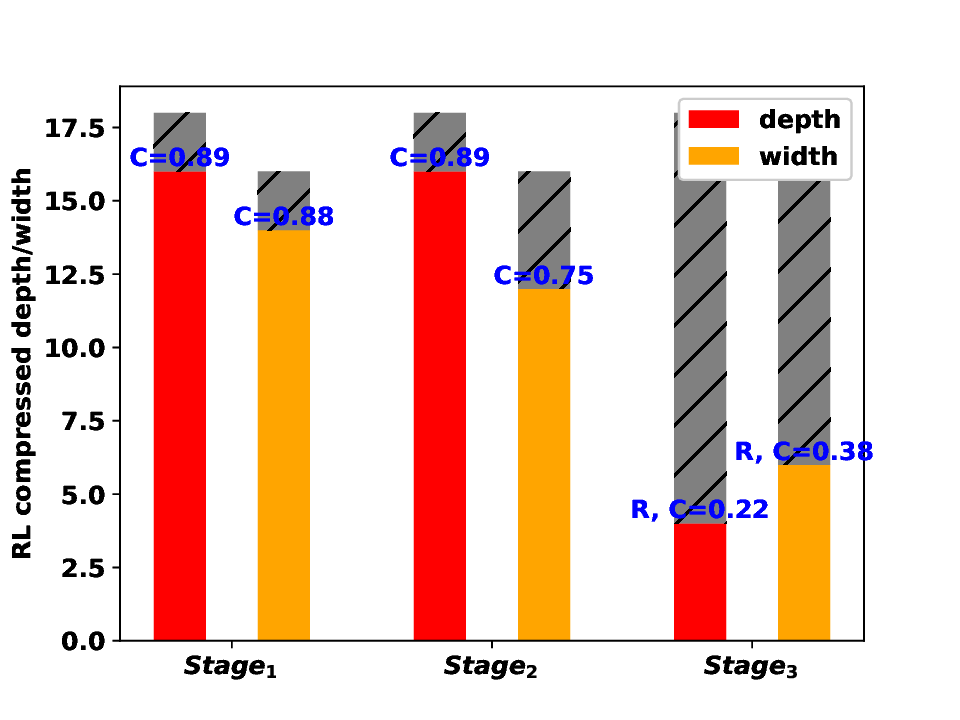}
      \caption{\label{fig:cifar100_20g_auto_rl} WRN-46-14 on auto-attack }
    \end{subfigure}
    \vspace{-2mm}
    \caption{Architecture topology analysis of \texttt{RobustResNet} in different attack scenarios from the mildest (\ie clean) to the most severe (\ie auto-attack) on CIFAR-10 (a)-(h) and CIFAR-100 (i)-(p) data sets. WRN-28-10, WRN-34-12 and WRN-46-14 are leveraged as teacher networks with 5, 10 and 20 GFLOPs computation budgets, respectively. The corresponding student networks are referred to as \texttt{RobustResNet-A1}, \texttt{RobustResNet-A2} and \texttt{RobustResNet-A3}. The entire teacher network architecture is partitioned into multiple (\eg 3) stages as in \cite{huang2022revisiting} and we visualize both depths and widths of the corresponding \texttt{RobustResNet} for each stage. In (a)-(d), the left three bar plots (in blue and orange) show the depths and widths in the three stages of \texttt{RobustResNet} A1 and A2 that follow the same configuration rules; the right three bar plots (in red and orange) show the adaptive configuration obtained by the proposed reinforced learning (RL) based architecture search. In (i)-(l), the left three bar plots represent depths and widths in the three stages of \texttt{RobustResNet} A2 and A3 and the right three bar plots show the RL compressed network's topology configuration.
    In (e)-(h) and (m)-(p), the grey bars denote the capacity of the corresponding teacher networks and $C$ denotes the remaining percentage of each stage after compression.
    }
    \label{fig:result_cifar}
    \vspace{-8mm}
\end{figure}

\subsection{Additional RL Critical Component Ablation Study}
\label{app:component_ablate}
We collect the additional RL designed component ablation study results on CIFAR-10 and CIFAR-100 under 10G computation budget from teacher network WRN-34-12 in Table~\ref{tab:ablate_component_cifar}.

\subsection{Additional Topology Comparison Results}
\label{app:topology}
We collect additional topology comparison results on diverse target task settings: (1) clean CIFAR-10 under 5G/10G budget, (2) auto attacked CIFAR-10 under 5G/10G budget, (3) clean CIFAR-100 under 10G/20G budget and (4) auto attacked CIFAR-100 under 10G/20G budget. The comparison results are shown in Figure~\ref{fig:result_cifar}.

\section{Broader Impact}
\label{app:impact}
Since nowadays data input from real environment is multi-modal, diverse and contain much potential unwanted noises, and most edge computing devices require an efficient memory usage of neural networks, shrinking the network parameter space without harming the generalization and adversarial robustness ability is tempting. Our method provides a novel reinforced compressive architecture search framework which could realize the difficulty level of the input scenarios and take corresponding compression operations to make the sampled neural network architecture more robust to the versatile input challenges after standard adversarial training.

\section{Source Code}
\label{app:code}
For the source code, please click \href{https://github.com/wdr123/robN2N}{here}.

\end{document}